\documentclass[11pt]{article}

\usepackage[final]{acl}
\usepackage{times}
\usepackage{latexsym}
\usepackage[T1]{fontenc}
\usepackage{microtype}
\usepackage{inconsolata}
\usepackage{graphicx}
\usepackage{booktabs}
\usepackage{amsmath,amssymb}
\usepackage{multirow}
\usepackage{arydshln}
\usepackage{tikz}
\usepackage{listings}
\usepackage{tcolorbox}
\usepackage{tabularx}
\usepackage{combelow}
\usetikzlibrary{positioning,arrows.meta,fit,backgrounds,calc,decorations.pathreplacing}

\newcommand{\Vocab}{\mathcal{V}^*}
\newcommand{\usefreq}{\hat\pi}

\usepackage[minimal]{leadsheets}
\useleadsheetslibraries{musicsymbols}

\usepackage{textcomp}

\newcommand{\smallclef}{%
  \raisebox{0.15ex}{\scalebox{0.55}{\trebleclef}}%
}

\title{DUET: Dual-MDL Unsupervised Entropy-Trie Segmentation \\ for Low-Resource Languages}

\author{
    Xiaotian Wang\,\smallclef\textsuperscript{1,3}
    \ \ \ \
    Youyuan Lin\,\smallclef\textsuperscript{2}
    \ \ \ \
    Zhan Shen\textsuperscript{1}
    \ \ \ \
    Hitomi Yanaka\textsuperscript{1,3,4}
    \\
    \textsuperscript{1}The University of Tokyo
    \ \
    \textsuperscript{2}Kyoto University
    \ \
    \textsuperscript{3}Riken
    \ \
    \textsuperscript{4}Tohoku University
    \\
    \textsuperscript{1}\texttt{\{xtwang, shenzhan, hyanaka\}@is.s.u-tokyo.ac.jp}
    \\
    \textsuperscript{2}\texttt{lin.youyuan.73v@st.kyoto-u.ac.jp}
}

\begin{document}

\maketitle

\begingroup
\makeatletter
\renewcommand{\@makefnmark}{%
  \hbox{%
    \raisebox{0.15ex}[0pt][0pt]{%
      \scalebox{0.55}{\trebleclef}%
    }%
    \hspace{0.2em}%
  }%
}
\footnotetext{: Equal contribution.}
\makeatother
\endgroup

\begin{abstract}
Morphological segmentation, the task of decomposing words into constituent morphemes, becomes particularly challenging under extreme data scarcity, a regime that can arise in language documentation and linguistic reasoning. This difficulty stems from existing approaches' reliance on annotated data, external resources, or recurring patterns that emerge only in larger corpora.
To address this limitation, we propose \textbf{D}ual-MDL \textbf{U}nsupervised \textbf{E}ntropy-\textbf{T}rie Segmentation~(\textbf{DUET}), a deterministic unsupervised algorithm that decouples structural candidate generation from statistical candidate ranking.
DUET combines branching entropy with recursive paradigm discovery to generate linguistically plausible candidate segmentations, then ranks them using a dual minimum description length~(MDL) criterion: a corpus-level objective rewards morpheme reuse, while a construction-tree objective guards against over-segmentation.
On the SIGMORPHON 2023 interlinear glossing shared task,
DUET outperforms unsupervised baselines in Morpheme F1 under both the standard and simulated extremely low-resource settings,
while providing broader candidate coverage.
Augmenting prompts with DUET segmentations improves performance on a linguistic reasoning benchmark, yielding the best overall results among the evaluated segmentation methods.
The code is publicly available at \url{https://github.com/ynklab/DUET}.

\end{abstract}

\section{Introduction}
\label{sec:intro}
    
Morphological segmentation, the task of identifying morpheme boundaries within words (e.g., segmenting \emph{walked} into \emph{walk} and \emph{-ed}), is fundamental to computational morphology~\citep{goldsmith-2001-unsupervised, cotterell-etal-2015-labeled, sigmorphon2022}. 
By revealing the internal structure of words, it facilitates language documentation~\citep{bird-2020-decolonising}, lexicon induction~\citep{irvine-callison-burch-2017-comprehensive}, and machine translation~\citep{belinkov-etal-2017-neural, toral-etal-2019-neural}. 
Recent neural approaches have substantially improved supervised morphological analysis, 
but require sufficient labeled data, limiting their applicability in extremely low-resource settings where the available data comprise only a few sentences or word types~\citep{sorokin-2019-convolutional, qi-etal-2020-stanza, rice-etal-2024-tams}.
Unsupervised alternatives avoid annotation but still depend on recurring patterns in the observed vocabulary. Representative approaches include compression-based methods~\citep{morfessor, morfessor2}, paradigm discovery~\citep{erdmann2020}, and Bayesian grammar induction~\citep{adaptorgrammar, morphagram}.

This reliance on recurring evidence is particularly problematic for endangered-language documentation and Linguistics Olympiad-style reasoning tasks, where lexical evidence is sparse~\citep{sahin-etal-2020-puzzling, lingoly, sigmorphon2023}. 
In such settings, the challenge is not merely limited sample size: a linguistically plausible segmentation may receive little or no support from other observed word forms. Consequently, methods that use corpus-wide recurrence to determine segmentation directly may favor fewer boundaries and thus under-segment.

To address this limitation, we propose \textbf{DUET} (\textbf{D}ual-MDL \textbf{U}nsupervised \textbf{E}ntropy-\textbf{T}rie Segmentation), a two-stage segmenter for extreme data scarcity that decouples structural candidate generation from statistical ranking.
The generation stage combines branching entropy over forward and backward tries with recursive paradigm discovery through maximal bicliques, producing structurally plausible candidate segmentations without global optimization.
The ranking stage then selects among these candidates using a dual-MDL criterion: a corpus-level objective rewards morpheme reuse, while a construction-tree objective guards against over-segmentation when corpus-level evidence is sparse. 
DUET is deterministic and requires no supervision, external resources, pretrained models, or language-specific tuning.
We evaluate DUET on the SIGMORPHON 2023 shared-task benchmark~\citep{sigmorphon2023} under both the standard settings and simulated extremely low-resource settings with as few as eight sentences.
We further evaluate its utility for downstream linguistic reasoning through segmentation-augmented prompting~\citep{sahin-etal-2020-puzzling}. Our contributions are summarized below:
\begin{itemize}
\vspace{-4pt}
\item

We introduce DUET, a deterministic two-stage framework for unsupervised morphological segmentation under extreme data scarcity, combining entropy-guided candidate generation with conservative dual-MDL ranking.

\vspace{-4pt}
\item

On SIGMORPHON 2023, DUET improves Morpheme F1, Oracle@1, and candidate coverage over unsupervised baselines across standard and simulated extremely low-resource settings;
on PuzzLing Machines, DUET-based prompting yields the strongest overall results among the evaluated segmentation methods.

\vspace{-4pt}
\item
We analyze the complementary roles of the two-stage MDLs through provenance, ablation, and matched-depth analyses, showing that the first provides stronger candidates while the second expands coverage.

\end{itemize}

\section{Related Work}
\label{sec:related}

\paragraph{Entropy-based segmentation.}
Branching statistics for morpheme boundary detection date back
to~\citet{harris1955}, who introduced successor variety
by counting the distinct phonemes following a prefix.
\citet{tanakaishii2005} recast this as branching
entropy, later used directly for boundary detection or to guide MDL
search~\citep{jin-tanaka-ishii-2006-unsupervised,
zhikov-etal-2010-efficient}.
DUET uses local entropy peaks as provisional cuts
for recursive paradigm discovery and reuses these
statistics as boundary evidence in ranking
(\S\ref{sec:method-gen}, \S\ref{sec:method-rank}).

\paragraph{Paradigm induction.}
Paradigm-based methods exploit recurring stem--affix co-occurrences.
Linguistica~\citep{goldsmith-2001-unsupervised} groups shared affixes
into \emph{signatures} and supports recursive stem analysis;
ParaMor~\citep{monson-etal-2007-paramor} searches stem--suffix schemes
and clusters partial paradigms; and
\citet{xu-etal-2018-unsupervised} refine probabilistic segmentations
with statistical paradigms.
DUET extracts maximal bicliques over prefix--stem--suffix frames and
recursively analyzes accepted chunks, pooling multi-depth analyses for
ranking (\S\ref{sec:method-gen}).

\paragraph{MDL-based segmentation.}
Morfessor~\citep{morfessor,morfessor2} and
Linguistica~\citep{goldsmith-2001-unsupervised,
lee-goldsmith-2016-linguistica} minimize a two-part
description length balancing model cost and corpus encoding cost.
For candidate selection, DUET uses a dual-MDL ranker:
a corpus-level objective rewards morpheme reuse, while a separate
construction-tree criterion is conditionally applied to discourage
over-segmentation under sparse evidence (\S\ref{sec:method-rank}).

\paragraph{Probabilistic segmentation.}
Adaptor Grammars perform nonparametric Bayesian
induction over reusable grammar subtrees~\citep{adaptorgrammar}
and underlie MorphAGram~\citep{morphagram}, a direct baseline.
Other probabilistic approaches include the log-linear model
of~\citet{poon2009}, 
combining overlapping features
with MDL-inspired priors, and the multilingual Bayesian model
of~\citet{snyder2008} over parallel text.
DUET performs deterministic segmentation from
monolingual word forms.

\section{Method}
\label{sec:method}

\begin{figure*}[!t]
\centering
    \includegraphics[
        width=\textwidth,
        trim=7.8cm 8.7cm 12.2cm 3.2cm,
        clip
        ]{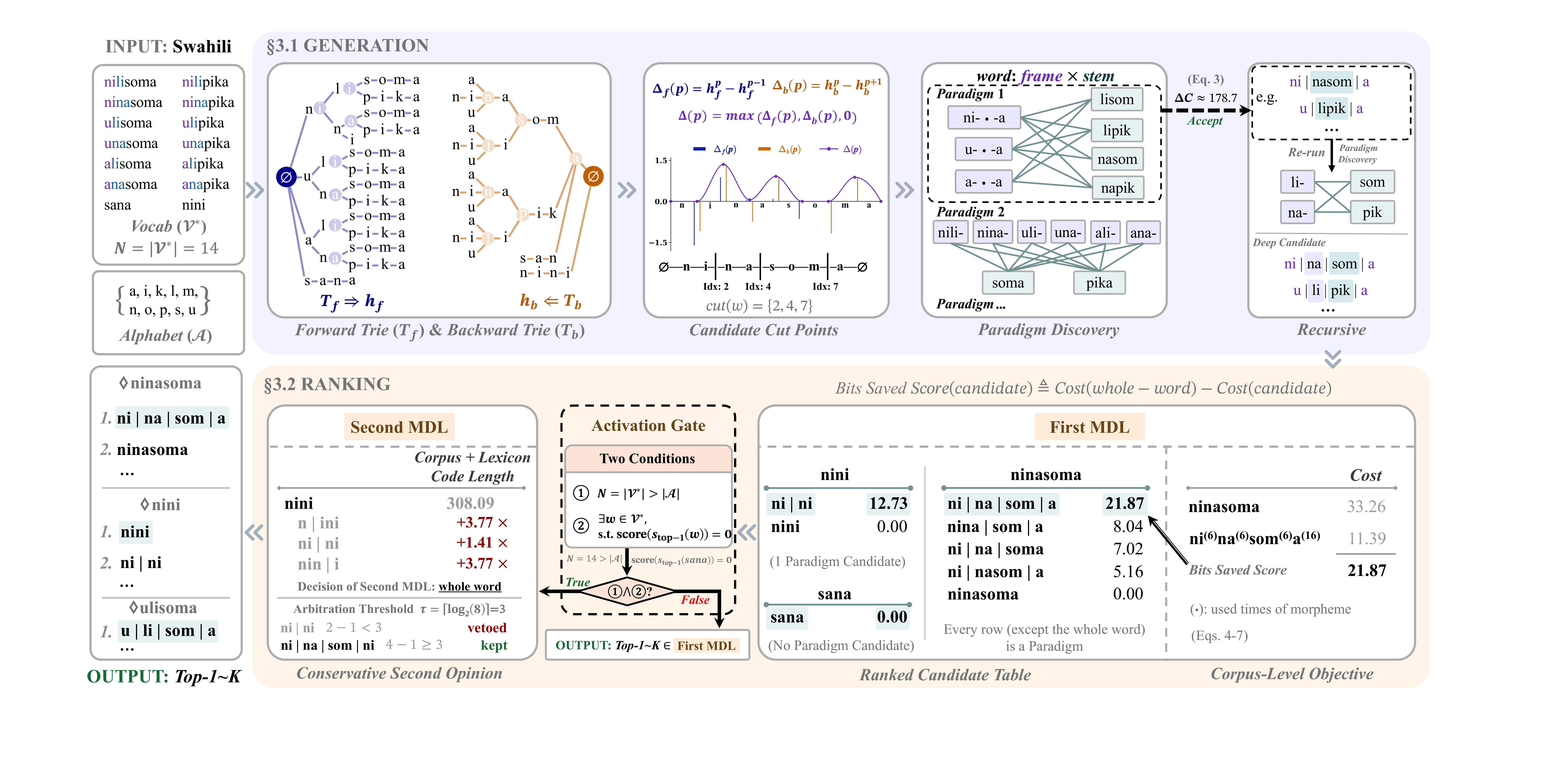}
\caption{
The entropy-trie dual-MDL framework illustrated with a running example: 
$N{=}14$ Swahili words (prefixes \texttt{ni-/u-/a-} $\times$ tense \texttt{-na-/-li-} $\times$ stems \texttt{soma}/\texttt{pika}, plus \texttt{sana} ``very'' and \texttt{nini} ``what'');
every number is the algorithm's actual output on this input.
A detailed description of the example is provided in Appendix~\ref{app:example}.
}
\label{fig:pipeline}
\vspace{-8pt}
\end{figure*}

This section details DUET, which consists of entropy-trie candidate generation (\S\ref{sec:method-gen}) followed by dual-MDL ranking (\S\ref{sec:method-rank}). 
Figure~\ref{fig:pipeline} illustrates the full pipeline using a running example.

\subsection{Candidate Generation}
\label{sec:method-gen}

An \emph{entropy-trie} combines forward and backward tries with branching-entropy estimates. The estimates are computed once and reused to propose cut points and recursively discover paradigms.

\subsubsection{Tries and Branching Entropy}

Let $\Vocab$ denote the word types in the input. Each word is normalized to Unicode NFC and tokenized into atoms, the smallest units between which boundaries may be placed; the distinct atoms form the alphabet $\mathcal A$. Each word and its reversal are inserted into a forward trie $T_f$ and a backward trie $T_b$, respectively. An end-of-word (EOW) symbol records whether a prefix is a complete word.
For a prefix $x$, \emph{branching entropy} quantifies uncertainty over the next atom. 
We estimate its distribution using additive-$\alpha$ smoothing with $\alpha=0.5$ (Jeffreys prior; \citealp{jeffreys1946}) over the observed continuations $\mathcal{C}(x)$ and an unseen-symbol event $\bot$:

\begin{equation}
\begin{aligned}
P(c\mid x)
&=\frac{\mathrm{count}(x\cdot c)+\alpha}
{\mathrm{count}(x)+\alpha\,(|\mathcal{C}(x)|+1)},\\
H(x)
&=-\!\!\sum_{c\in\mathcal{C}(x)\cup\{\bot\}}\!\!
P(c\mid x)\log_2 P(c\mid x).
\end{aligned}
\label{eq:H}
\end{equation}
For boundary position $p$ in word $w$, $h_f^p=H(w_{[:p]})$ from $T_f$ measures uncertainty over the following atom, while $h_b^p$ from $T_b$ analogously measures uncertainty over the preceding atom.

\subsubsection{Candidate Cut Points}

An increase in branching entropy across a position signals a potential morpheme boundary.
With entropy increases $\Delta_f(p)=h_f^{p}-h_f^{p-1}$, $\Delta_b(p)=h_b^{p}-h_b^{p+1}$, define boundary strength $\Delta(p)=\max(\Delta_f(p),\Delta_b(p),0)$.
A word's length is $n=|w|$~(in atoms),
and the same notation gives a morpheme's length (e.g., $|m_j|$ in Eq.~\ref{eq:corpusMDL}).
Candidate cut points are the \emph{positive local maxima} of $\Delta$:
\begin{gather}
\mathrm{cuts}(w)=\{p\in\{1,\ldots,n{-}1\}:\Delta(p)>0,\notag\\
\Delta(p)\ge\Delta(p{-}1),\ \Delta(p)\ge\Delta(p{+}1)\},
\label{eq:cuts}
\end{gather}
where out-of-range neighbors have strength $0$.

\subsubsection{Paradigm Discovery}
For each word $w$, every pair $i<j$ in $\{0,n\}\cup\mathrm{cuts}(w)$ defines a candidate split $(L,S,R)$ = (left affix, stem, right affix). 
We construct a bipartite graph with frame nodes $(L,R)$ and stem nodes $S$, adding an edge whenever $LSR\in\Vocab$. 
A \emph{paradigm} is an inclusion-maximal set of frames and stems whose Cartesian product is fully attested in $\Vocab$~\citep{erdmann2020}. 
For each stem pair, we collect their shared frames and every stem connected to all of them, 
forming a maximal biclique; 
we repeat this symmetrically for each frame pair.
We retain a paradigm only if it yields a shorter encoding of the words it covers. 
For the frame set $\mathcal F$, stem set $\mathcal S$, 
and $n_w=|\mathcal F||\mathcal S|$ covered word forms,

\begin{align}
C_{\text{base}} &= n_w\log_2 N\notag\\
&\quad+\sum_{(L,R)\in\mathcal F}\sum_{S\in\mathcal S}(|L|{+}|S|{+}|R|)\log_2|\mathcal A|,\notag\\
C_{\text{model}}&= \sum_{(L,R)\in\mathcal F}\big((|L|{+}|R|)\log_2|\mathcal A|+\log_2 N\big)\notag\\
&\quad+\sum_{S\in\mathcal S}\big(|S|\log_2|\mathcal A|+\log_2 N\big),\notag\\
C_{\text{data}} &= n_w\,(\log_2|\mathcal F|+\log_2|\mathcal S|),\notag\\
\Delta C &= C_{\text{base}}-C_{\text{model}}-C_{\text{data}}.
\label{eq:paradigmMDL}
\end{align}
% where $N=|\Vocab|$ and $C$ denotes a code length in bits. $C_{\text{base}}$ is the cost of spelling out $n_w$ words letter by letter; $C_{\text{model}}+C_{\text{data}}$ is the cost of storing two small tables plus one (frame, stem) index per word.
% A paradigm is kept only if $\Delta C\ge1$ bit: one full bit of evidence.
% Every $(L,S,R)$ split is only a \emph{candidate}: leaving a word whole remains an option throughout.
Here, $N=|\Vocab|$, and all costs are measured in bits. $C_{\text{base}}$ encodes each covered word independently, whereas $C_{\text{model}}+C_{\text{data}}$ encodes each frame and stem once, plus one frame-stem index per word. We retain paradigms with $\Delta C\ge1$ bit. All induced splits remain candidates, and the whole-word analysis is always available.

\subsubsection{Recursive Affix Unpacking}

% A single $(L,S,R)$ triple only cuts twice,
% missing agglutinative chains like the running example's \texttt{ni-na-som-a} (``I-PRES-read-FV''): a word-level split reaches \texttt{ni\,|\,nasom\,|\,a} but not all three internal boundaries.

A single $(L,S,R)$ split introduces at most two boundaries and thus cannot recover all boundaries, as illustrated by the running example \texttt{ni-na-som-a} (``I-PRES-read-FV''): the word-level split \texttt{ni\,|\,nasom\,|\,a} leaves the internal boundary in \texttt{nasom} unresolved. We therefore recursively apply biclique discovery to accepted blocks of length at least two using the existing candidate cuts; splitting \texttt{nasom} into \texttt{na}+\texttt{som} yields the full candidate \texttt{ni\,|\,na\,|\,som\,|\,a} (Figure~\ref{fig:pipeline}). 
Because every non-trivial split strictly shortens a block, the recursion terminates.

\subsection{Ranking: Dual-MDL Selection}
\label{sec:method-rank}

DUET ranks the generated candidates in two stages: an MDL-based scoring rule first rewards reusable segments, 
after which a conservative construction-tree criterion checks analyses that may over-segment around spuriously recurring substrings under sparse evidence.

\subsubsection{The First MDL}
The first MDL scores structural candidates using corpus-wide morpheme-usage statistics and boundary evidence. Because these statistics depend on the segmentation selected for every word, we seek a \emph{self-consistent} joint assignment in which each word selects a lowest-cost candidate under the induced statistics.
The search considers only structural candidates: the whole-word baseline and paradigm-licensed splits.
Let $\mathrm{use}(m)$ denote the number of occurrences of morpheme $m$
in the current segmentation assignment, let $M=\sum_{m'}\mathrm{use}(m')$, and let $\mathcal D=\{m:\mathrm{use}(m)>0\}$ be the set of morphemes used.
We update $M$ and $\mathcal D$ whenever a word's segmentation changes.
Using the same Jeffreys $\tfrac12$-smoothing as in Eq.~\ref{eq:H}, we define the \emph{usage frequency} of $m$ as
\begin{equation}
\usefreq(m)=
\frac{\mathrm{use}(m)+\frac12}
{M+\frac12(|\mathcal D|+1)}.
\label{eq:usagefreq}
\end{equation}
For a split $s=(m_1,\ldots,m_k)$ using cut positions $\mathrm{cut}(s)$ (a subset of the word's own candidate cuts $\mathrm{cuts}(w)$ from Eq.~\ref{eq:cuts}), we take the corpus-wide mean boundary strength over all candidate word--position pairs:
\begin{align}
\mathcal B&=\bigcup_{w\in\mathcal V}\{w\}\mathbin{\times}\mathrm{cuts}(w), \notag\\
\mu&=\frac{1}{|\mathcal B|}\sum_{(w,p)\in\mathcal B}\Delta_w(p).
\label{eq:meanboundary}
\end{align}
When $\mathcal B$ is empty, we set $\mu=0$.
Its cost is
\begin{gather}
\mathrm{cost}(w\leftarrow m_1\cdots m_k) = \notag\\
\underbrace{\sum_{j=1}^{k}\Big[-\log_2\usefreq(m_j)+\mathbf 1[m_j\notin\mathcal D]\,|m_j|\log_2|\mathcal A|\Big]}_{\text{usage cost}}\notag\\
{}-\underbrace{\sum_{p\in\mathrm{cut}(s)}\big(\Delta(p)-\mu\big)}_{\text{boundary-evidence bonus}}.
\label{eq:corpusMDL}
\end{gather}
The \emph{usage cost} assigns lower cost to frequently reused morphemes, while newly introduced morphemes incur an additional form-encoding cost, discouraging one-off fragments. The \emph{boundary-evidence bonus} lowers the cost of cuts with above-average boundary strength and raises that of weaker cuts~\citep{magistry2012}. The two terms are complementary: usage alone may favor frequent but spurious substrings, whereas boundary evidence alone may miss recurring morphemes with weak entropy signals.
Additionally, 
we discuss experiments introducing a weighting parameter $\lambda$ into Eq.~\ref{eq:corpusMDL} to balance the two terms in Appendix~\ref{app:eq6_lambda}.

We use a deterministic MDL-guided search,
initializing each word with its finest structural candidate to expose shared morphemes from the outset and then repeatedly updating words in a fixed order to their lowest-cost candidates under the current usage statistics until no assignment changes.
After the search,
we score each candidate using leave-one-out statistics: before evaluating the candidates for word $w$, we temporarily remove the morphemes in its current segmentation from $\mathrm{use}(\cdot)$ and $\mathcal D$, then restore them afterward. This prevents $w$ from contributing to its own usage-frequency estimates.
The score of candidate segmentation $s$ is the number of bits saved relative to the whole-word analysis:
\begin{equation}
\mathrm{score}(s)=\mathrm{cost}(\text{whole-word})-\mathrm{cost}(s).
\label{eq:score}
\end{equation}
The whole-word baseline scores 0. 
Candidates are ranked by decreasing score, with zero-score ties resolved in favor of the baseline; each entry retains its structural derivation.

\subsubsection{The Second MDL}
To prevent over-segmentation from spuriously recurring substrings,
we introduce a more conservative description-length objective drawing on Morfessor~\citep{morfessor, morfessor2, smit-etal-2014-morfessor}.
It uses a recursive binary construction tree $\theta$ that begins with whole words 
and accepts a split only when it reduces the joint corpus-and-lexicon code length.
For a count vector $\mathbf{n}=(n_1,\ldots,n_k)$, let
$n=\sum_{i=1}^{k}n_i$ and let
$\widehat{\mathbf{p}}_{\mathbf{n}}$ denote its empirical
distribution, with $\widehat{p}_i=n_i/n$. We define
\begin{equation}
\mathcal{L}_0(\mathbf{n})
=
nH_0\!\left(\widehat{\mathbf{p}}_{\mathbf{n}}\right)
=
n\log_2 n
-
\sum_{i=1}^{k}n_i\log_2 n_i,
\label{eq:zero-order-code-length}
\end{equation}
Here, $H_0(\widehat{\mathbf{p}}_{\mathbf{n}})$ is the zeroth-order empirical entropy in bits per symbol, and $\mathcal{L}_0(\mathbf{n})$ is the corresponding total code length~\citep{manzini2001analysis}.
The joint corpus-and-lexicon code length is defined as follows:
\begin{equation}
\begin{gathered}
\begin{aligned}
\mathbf{c}_{\theta}
&=
\left(
  (c_m)_{m\in\mathcal{D}_{\theta}},
  B_C
\right),
&
T
&=
\sum_{m\in\mathcal{D}_{\theta}} c_m,
\\
\mathbf{f}_{\theta}
&=
\left(
  (f_a)_{a\in\mathcal{A}_{\theta}},
  |\mathcal{D}_{\theta}|
\right),
&
N_L
&=
\sum_{a\in\mathcal{A}_{\theta}} f_a
+
|\mathcal{D}_{\theta}|,
\end{aligned}
\\[3mm]
\begin{aligned}
L(\theta)
&=
\underbrace{
  \mathcal{L}_0(\mathbf{c}_{\theta})
  +
  \log_2
  \tbinom{T-1}{|\mathcal{D}_{\theta}|-1}
}_{L_{\mathrm{corpus}}(\theta)}
\\[2mm]
&\quad+
\underbrace{
  \mathcal{L}_0(\mathbf{f}_{\theta})
  +
  \log_2
  \tbinom{N_L-1}{|\mathcal{A}_{\theta}|}
  -
  \log_2\!\left(|\mathcal{D}_{\theta}|!\right)
}_{L_{\mathrm{lexicon}}(\theta)} .
\end{aligned}
\end{gathered}
\label{eq:second-mdl}
\end{equation}
Here, $\mathcal{D}_{\theta}$ denotes the leaf-morpheme lexicon of tree $\theta$. 
$c_m$ and $T$ are the frequency of morpheme $m$
and the total number of morpheme occurrences, respectively,
while $B_C$ is the
word-boundary count. Likewise, $\mathcal{A}_{\theta}$ is the atom
inventory, $f_a$ is the lexicon frequency of atom $a$, and $N_L$ is
the total number of atom and lexicon-entry-boundary tokens.

Since the second MDL does not use paradigm information, it provides an independent check based solely on compression evidence.
Writing $s_{\mathrm{MDL\text{-}1}}^{r_1}(w)$ for the top-ranked first-MDL segmentation of word $w$, we activate the second MDL only when
\begin{equation}
% \text{tree active}
% \iff
N>|\mathcal A|
\ \land\
\exists w:\mathrm{score}(s_{\mathrm{MDL\text{-}1}}^{r_1}(w))=0.
\label{eq:treeactive}
\end{equation}
The second MDL is therefore skipped when $N\le|\mathcal A|$ or when every top-ranked first-MDL candidate has positive score.
When the second MDL is active, we set the arbitration threshold\footnote{Threshold selection is discussed in Appendix~\ref{app:tau-ablation}.} to $\tau=\lceil\log_2(\tilde n+1)\rceil$, where $\tilde n$ is the median word length in $\Vocab$.
A non-positive top-1 has not beaten the whole word in the first MDL and is left unchanged.
For each remaining word $w$, we select its final top-ranked segmentation $s^{r_1}$ as follows:
\begin{equation}
s^{r_1} = 
\begin{cases}
s_{\mathrm{MDL\text{-}1}}^{r_1}, & \text{if } |s_{\mathrm{MDL\text{-}1}}^{r_1}| - |s_{\mathrm{MDL\text{-}2}}^{r_1}| \ge \tau, \\
s_{\mathrm{MDL\text{-}2}}^{r_1}, & \text{otherwise,}
\end{cases}
\label{eq:secondopinion}
\end{equation}
where $|\cdot|$ denotes the number of morphemes.

After selecting the final top-ranked analysis, we merge the remaining candidates by rank.
For ranker $j\in{\mathrm{MDL\text{-}1},\mathrm{MDL\text{-}2}}$, let $C_j$ denote its candidate list and $r_j(s)$ the rank of candidate $s$. We assign $s$ the merged rank mass
\begin{equation}
q(s)=\frac12\sum_j
\frac{\mathbf 1[s\in C_j]}{r_j(s)(r_j(s)+1)}.
\label{eq:rankmixture}
\end{equation}

\vspace{-4pt}

\noindent Since $\sum_{r\ge1}1/[r(r+1)]=1$, $q$ is an equal-weight mixture of two proper rank distributions,
assigning mass from both lists to shared candidates.
We sort by decreasing $q$, remove duplicates, and reserve the last slot for the whole-word baseline if needed.

\begin{figure*}[htbp]
\centering
\includegraphics[
        width=\textwidth,
        trim=0.0cm 0.38cm 0.0cm 0.0cm,
        clip
        ]{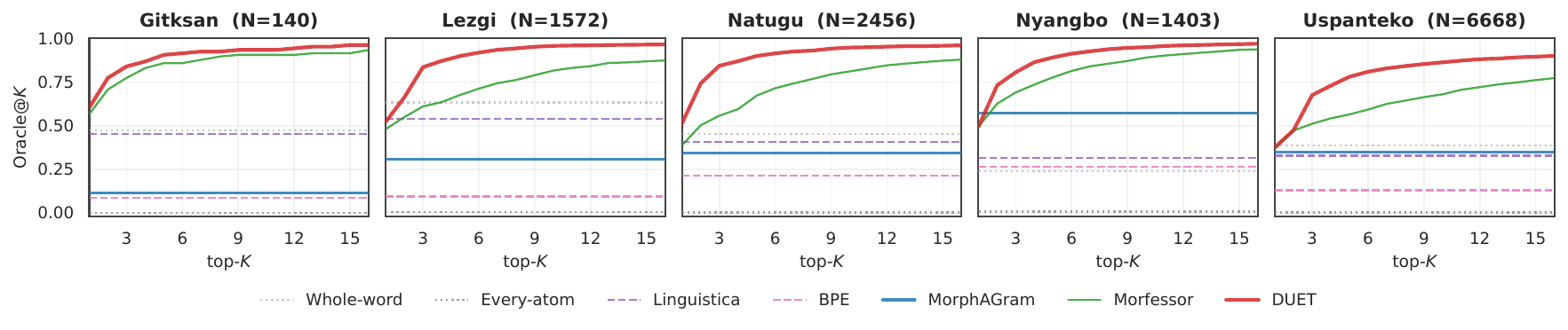}
\caption{Oracle@$K$ by final candidate budget.
Single-output systems are horizontal at Oracle@1.}
\label{fig:oracle-train}
\vspace{-8pt}
\end{figure*}

\vspace{-10pt}
\section{Experimental Setup}
\label{sec:exp}

\paragraph{Compared Systems.} Two trivial bounds frame the range: \texttt{whole-word} (never split) and \texttt{every-atom} (split everywhere).
We compare against three classic unsupervised segmenters: 
\textbf{Morfessor 2.0}~\citep{morfessor2, smit-etal-2014-morfessor} 
(MDL-based), 
\textbf{Linguistica}~\citep{goldsmith-2001-unsupervised, lee-goldsmith-2016-linguistica}
(shared-affix parsing; we use only its classic suffixal
right-to-left parse, which produces one deterministic analysis per word),
and \textbf{MorphAGram}~\citep{morphagram}.
We also include \textbf{BPE}~\citep{sennrich2016} as a general-purpose reference, applying merges to the same atom sequences and selecting the merge count with an MDL objective
(see Appendix~\ref{app:implementation} for details).
We report runtime comparisons and analyze algorithmic complexity in Appendix~\ref{app:complexity}.

\paragraph{Data.} We use five languages from the SIGMORPHON 2023 interlinear glossing shared task~\citep{sigmorphon2023},
which provides gold morphological segmentations for each word. 
However, the gold segmentation may be \emph{canonical},
recovering underlying morpheme forms obscured by phonological alternations that are not transparently reflected in the orthography; consequently, the gold morphemes do not always concatenate to the surface word. 
By contrast, DUET and the comparison systems produce only \emph{surface} segmentations. 
We therefore evaluate on the \emph{Surface Subset}, comprising words whose normalized gold morphemes concatenate exactly to the surface form. Table~\ref{tab:data} summarizes the dataset statistics. 
We report results on the training split unless otherwise noted, with development and test results presented in Appendix~\ref{app:devtest}.
Additionally, we discuss experimental results 
on the SIGMORPHON 2022 shared task~\citep{sigmorphon2022} in Appendix~\ref{app:blind2022}.

\begin{table}[htbp]
\centering
\resizebox{\columnwidth}{!}{
\begin{tabular}{lrrrr}
\toprule
Language & Sentences & Word types & Surface & Cover. \\
\midrule
Gitksan   & 31   & 140  & 106  & 76\% \\
Lezgi     & 634  & 1572 & 983  & 63\% \\
Natugu    & 789  & 2456 & 1662 & 68\% \\
Nyangbo   & 1670 & 1403 & 1316 & 94\% \\
Uspanteko & 9116 & 6668 & 5111 & 77\% \\
\bottomrule
\end{tabular}
}
\caption{Statistics of the surface subset in the SIGMORPHON 2023 training set.
Coverage is the surface-subset fraction of all word types.}
\label{tab:data}
\end{table}

\paragraph{Metrics.} 
We evaluate performance using three metrics that expose different failure modes.
\textbf{Morpheme F1}\footnote{
It is the official SIGMORPHON 2022 metric.
} 
\citep{sigmorphon2022} matches predicted and gold morpheme sequences via the longest common subsequence and micro-averages the F1;
it penalizes over-segmentation directly, 
since every spurious extra cut inserts an unmatched item.
\textbf{Boundary F1} \citep{virpioja2011} scores predicted vs. gold internal cut points, 
micro-averaged over all boundary positions
(single-morpheme words have none and are excluded); 
it is less sensitive to over-segmentation than Morpheme F1 as it considers only boundary placement.
\textbf{Oracle@}$\boldsymbol{K}$ scores a ranked candidate list: a hit if any of the top $K$ candidates reproduces the \emph{entire} gold morpheme sequence exactly.
When $K=1$, this reduces to the exact-match accuracy
(\textbf{EM}) of the top-ranked candidate.
By definition, Oracle@1 $\le$ Oracle@$K$.
The gap between Oracle@$K$ and Oracle@1 indicates how often a correct segmentation appears among the top $K$ candidates but is not ranked first.
Every system that can produce a ranked list is scored up to $K{=}16$.

\begin{figure*}[t]
\centering
\includegraphics[
        width=\textwidth,
        trim=0.0cm 0.38cm 0.0cm 0.0cm,
        clip
        ]{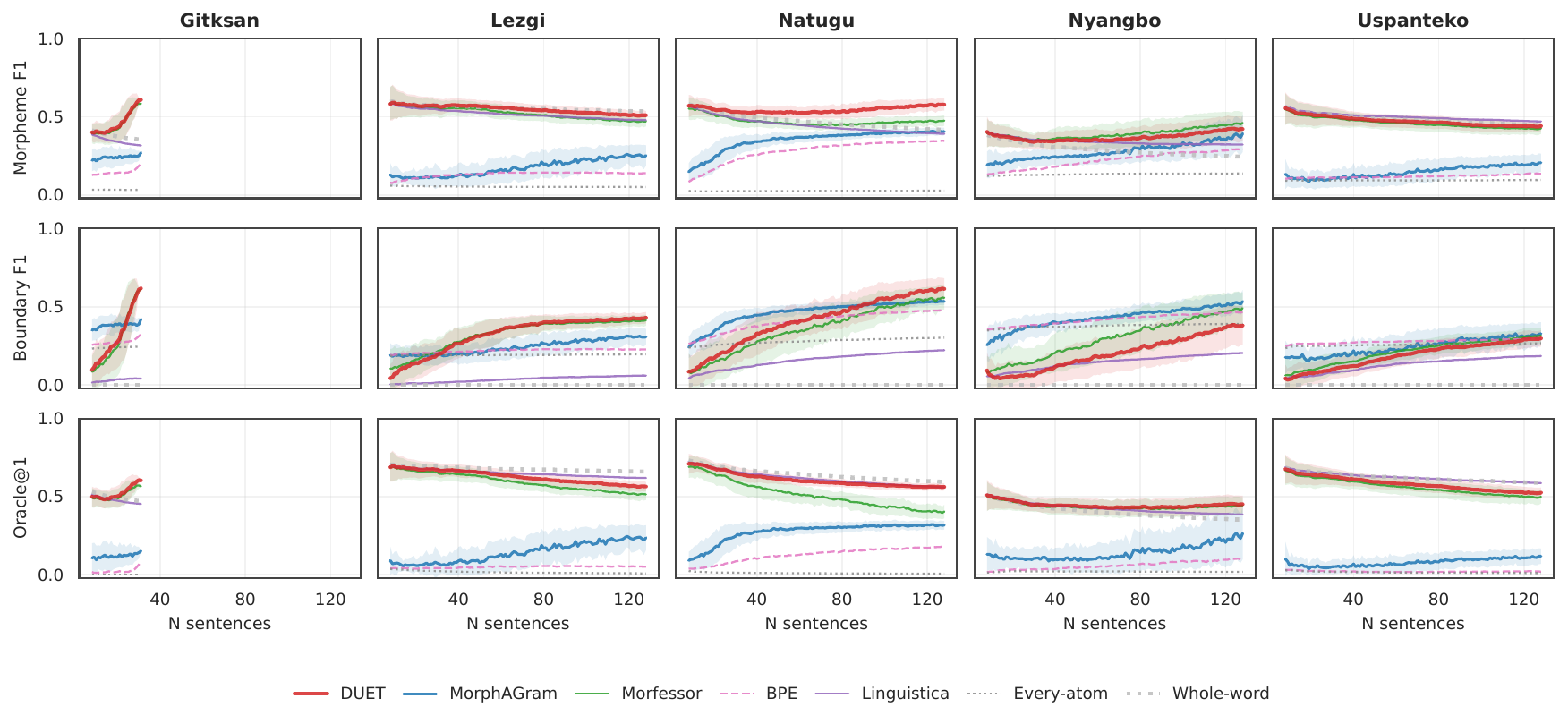}
\caption{Extremely low-resource learning curves (5-language sentence-sampling means, raw per-$N$ values, $\pm1$ SD shading over 64 seeds).
Top to bottom: Morpheme F1, Boundary F1, Oracle@1.}
\label{fig:lowres-curves}
\vspace{-8pt}
\end{figure*}

\section{Results}
\label{sec:results}

\subsection{Main Experimental Results}
\label{sec:train-results}

Table~\ref{tab:train-results} reports macro (equal-weight across languages) and weighted (by surface-subset size) averages for all three metrics; full per-language numbers for all three splits are in Appendix~\ref{app:devtest}.

\begin{table}[htbp]
\centering
\small
\setlength{\tabcolsep}{3.2pt}
\resizebox{\columnwidth}{!}{
\begin{tabular}{lrrrrrr}
\toprule
 & \multicolumn{2}{c}{Morph.\ F1} & \multicolumn{2}{c}{Bnd.\ F1} & \multicolumn{2}{c}{Oracle@1} \\
\cmidrule(lr){2-3}\cmidrule(lr){4-5}\cmidrule(lr){6-7}
System & macro & wtd. & macro & wtd. & macro & wtd. \\
\midrule
whole-word  & .309 & .272 & .000 & .000 & .437 & .405 \\
every-atom  & .074 & .092 & .312 & .334 & .004 & .004 \\
\midrule
BPE         & .296 & .316 & .430 & .458 & .157 & .160 \\
Linguistica & .323 & .319 & .239 & .330 & .408 & .364 \\
% Ent-MDL     & .429 & .490 & .506 & .568 & .362 & .428 \\
MorphAGram  & .433 & .468 & .548 & .588 & .337 & .372 \\
Morfessor   & .519 & .487 & .597 & \textbf{.591} & .460 & .406 \\
DUET        & \textbf{.541} & \textbf{.500} & \textbf{.601} & \textbf{.591} & \textbf{.502} & \textbf{.436} \\
\bottomrule
\end{tabular}
}
\caption{Results on the surface subset of SIGMORPHON 2023 training set.
\textbf{Bold} is best among the five non-trivial systems.}
\label{tab:train-results}
\end{table}

On Morpheme F1 and Oracle@1, the two metrics most sensitive to over-segmentation, 
DUET has the widest margin over Morfessor (macro .541 vs.\ .519 and .502 vs.\ .460); on Boundary F1,
the two are essentially tied (macro .601 vs.\ .597, weighted .591 vs.\ .591).
BPE trails every morphology-aware system on all three metrics despite using the same MDL-style automatic stopping criterion, 
suggesting that compressing atom sequences 
differs from finding morphological structure.
The Oracle@$K$ results are presented in Figure~\ref{fig:oracle-train}.
DUET consistently outperforms Morfessor's $K$-best Viterbi list at every Oracle@$K$,
with Oracle@16 gains ranging from Gitksan (.972 vs. .934) to Uspanteko (.898 vs. .773).

\subsection{Extremely Low-Resource Settings}
\label{sec:lowres}

To simulate an extremely low-resource setting,
we randomly sample $N\in\{8,\dots,128\}$ sentences 
(with $N{=}8$ corresponding to 30--66 word types per language) from
each language, repeat 64 times, 
and average every metric over the seeds.

\begin{table}[htbp]
\centering
% \small
\setlength{\tabcolsep}{4pt}
\resizebox{\columnwidth}{!}{
\begin{tabular}{lrrrr}
\toprule
System & Morph.\ F1 & Bnd.\ F1 & Oracle@1 & Oracle@16 \\
\midrule
whole-word  & .448 & .000 & .582 & -- \\
every-atom  & .075 & .275 & .015 & -- \\
\midrule
BPE         & .187 & .328 & .063 & -- \\
Linguistica & .447 & .113 & \textbf{.574} & -- \\
% Ent-MDL     & .098 & .288 & .028 & -- \\
MorphAGram  & .239 & \textbf{.347} & .157 & -- \\
Morfessor   & .463 & .299 & .526 & .938 \\
DUET        & \textbf{.484} & .282 & .564 & \textbf{.950} \\
\bottomrule
\end{tabular}
}
\caption{
Results on the extremely low-resource setting ($N{=}8$--$128$, 5 languages $\times$ 64 seeds).
\textbf{Bold} is best among the five non-trivial systems; systems without an $n$-best list have no Oracle@16 entry (marked ``--'').}
\label{tab:lowres-main}
\end{table}

As shown in Table~\ref{tab:lowres-main},
DUET achieves the best Morpheme F1 (.484) and Oracle@16 (.950).
Its Oracle@1 (.564) nears Linguistica's .574, 
while its Boundary F1 (.282) trails Morfessor (.299) and MorphAGram (.347).
Figure~\ref{fig:lowres-curves} makes this trade-off explicit.
BPE again performs poorly:
although its Boundary F1 (.328) approaches MorphAGram's, 
its Morpheme F1 (.187) and Oracle@1 (.063) are the worst among non-trivial systems, 
reinforcing that generic compression is no substitute for morphology-specific inductive bias.
We also audit whether a zero-shot LLM can perform this segmentation itself
under the same character-blind protocol;
the prompt, settings, and results are reported in Appendix~\ref{app:llm-audit}.

\begin{table*}[t]
\centering
\resizebox{\textwidth}{!}{
\begin{tabular}{lcccccccc}
\toprule
\multirow{2}{*}{\textbf{Methods}}
& \multicolumn{4}{c}{\textbf{Original}}
& \multicolumn{4}{c}{\textbf{Pseudo}} \\
\cmidrule(lr){2-5} \cmidrule(lr){6-9} 
& \textbf{BLEU-2} & \textbf{chrF-3} & \textbf{cTER} & \textbf{EM}
& \textbf{BLEU-2} & \textbf{chrF-3} & \textbf{cTER} & \textbf{EM} \\
\midrule
\textsc{Qwen3.6}: ICL& $68.50\pm1.01$ & $85.31\pm0.26$ & $85.19\pm0.29$ & $53.92\pm1.39$ & $55.86\pm1.16$ & $75.92\pm0.62$ & $74.79\pm0.79$ & $38.27\pm1.00$ \\
\quad + LLMSeg 
& $69.51\pm1.91$ & $85.94\pm0.79$ & $85.92\pm1.21$ & $54.81\pm2.32$ & $56.34\pm1.09$ & $75.73\pm0.64$ & $74.47\pm0.81$ & $38.49\pm1.08$ \\
\quad + Morfessor 
& $70.06\pm1.14$ & $86.05\pm0.33$ & $85.95\pm0.93$ & $55.43\pm1.64$ & $56.48\pm0.45$ & $75.99\pm0.31$ & $74.55\pm0.79$ & $38.67\pm0.55$   \\
\quad + DUET   
& $\textbf{70.83}\pm0.94$ & $\textbf{86.10}\pm0.44$ & $\textbf{86.11}\pm0.58$ & $\textbf{56.85}\pm1.08$ & $\textbf{57.73}\pm0.33$ & $\textbf{76.63}\pm0.55$ & $\textbf{74.97}\pm0.78$ & $\textbf{40.46}\pm0.50$ \\
\cdashline{1-9}
\textsc{Gemma4}: ICL 
& $77.81\pm0.93$ & $90.28\pm0.60$ & $91.13\pm0.70$ & $66.94\pm1.44$ & $63.72\pm2.48$ & $81.73\pm0.91$ & $81.06\pm1.10$ & $47.38\pm2.78$\\
\quad + LLMSeg 
& $78.15\pm0.65$ & $90.13\pm0.60$ & $90.75\pm0.71$ & $67.47\pm1.18$ & $65.94\pm1.03$ & $\textbf{82.21}\pm0.79$ & $\textbf{81.69}\pm0.58$ & $50.19\pm1.33$ \\
\quad + Morfessor 
& $78.37\pm1.29$ & $90.03\pm0.92$ & $90.63\pm1.01$ & $66.98\pm1.35$ & $63.84\pm1.89$ & $80.77\pm1.23$ & $79.91\pm1.14$ & $48.86\pm1.77$\\
\quad + DUET 
& $\textbf{78.69}\pm0.43$ & $\textbf{90.50}\pm0.38$ & $\textbf{91.27}\pm0.40$ & $\textbf{67.62}\pm0.57$ & $\textbf{66.35}\pm0.65$ & $82.12\pm0.26$ & $81.61\pm0.39$ & $\textbf{51.11}\pm1.08$\\
\cdashline{1-9}
\textsc{GPT-oss}: ICL            
& $66.49\pm1.05$ & $82.62\pm0.86$ & $81.66\pm1.14$ & $51.33\pm1.43$ & $47.06\pm1.95$ & $67.93\pm1.43$ & $65.17\pm1.73$ & $28.80\pm2.63$\\
\quad + LLMSeg   
& $65.00\pm1.76$ & $82.36\pm0.72$ & $81.09\pm0.98$ & $49.57\pm2.00$ & $48.02\pm1.39$ & $\textbf{68.35}\pm1.40$ & $\textbf{65.69}\pm1.01$ & $29.91\pm1.93$\\
\quad + Morfessor 
& $65.45\pm1.43$ & $82.24\pm0.88$ & $81.25\pm1.30$ & $49.69\pm1.94$ & $48.25\pm2.70$ & $68.12\pm1.52$ & $65.09\pm1.68$ & $29.88\pm3.19$\\
\quad + DUET      
& $\textbf{66.70}\pm0.60$ & $\textbf{82.87}\pm1.23$ & $\textbf{82.02}\pm1.28$ & $\textbf{51.42}\pm0.58$ & $\textbf{48.59}\pm1.38$ & $67.96\pm1.29$ & $64.88\pm1.22$ & $\textbf{30.43}\pm1.72$\\
\bottomrule
\end{tabular}
}
\caption{
Results on the PuzzLing Machines Benchmark. 
Each method is evaluated over 5 runs and reported as mean $\pm$ standard deviation.
The best result for each metric under each backbone model is shown in \textbf{bold}.
}
\label{tab:puzzle_results}
\end{table*}

\subsection{Downstream Linguistic Reasoning}
\label{sec:ds_eval}

Linguistic puzzle benchmarks evaluate whether LLMs can infer grammatical rules from a few examples and apply them to solve new problems.
These examples typically consist of sentences in a low-resource language, sometimes accompanied by translations~\citep{sahin-etal-2020-puzzling, linguini, lingoly, khouja2026lingolytoo}.
To evaluate whether explicit segmentation aids reasoning, we add segmentations of all low-resource-language words to the in-context learning (ICL)\footnote{
Standard ICL that directly prompts the model with the context and the question, 
serving as the primary baseline~\citep{lin-etal-2023-solving, lin2025probinglargelanguagemodels, ramji-ramji-2025-inductive}.
}
prompt. 
We evaluate on the PuzzLing Machines Benchmark~\citep{sahin-etal-2020-puzzling} using its official script and four metrics: BLEU-2~\citep{papineni-etal-2002-bleu}, chrF-3~\citep{popovic-2015-chrf}, cTER~\citep{wang-etal-2016-character}, and exact match (EM).
We compare DUET with Morfessor and LLMSeg,
where the backbone LLM generates segmentation directly.
For each word, up to eight candidates are generated 
using \textsc{Qwen3.6-27B-FP8}~\citep{qwen3.6-27b},
\textsc{Gemma4-31B-it}~\citep{gemmateam2026gemma4},
and \textsc{GPT-oss-120B}~\citep{openai2025gptoss120bgptoss20bmodel}. 
To reduce potential lexical leakage, we create a pseudo benchmark by applying a deterministic one-to-one character substitution to all low-resource-language words. Details of the mapping, experimental setup, and prompts appear in Appendix~\ref{app:puzzling_machines}.

Table~\ref{tab:puzzle_results} summarizes the results.
LLMSeg often matches or outperforms Morfessor, making it a competitive baseline. 
DUET achieves the best results across most combinations of backbone models and evaluation metrics. 
Its improvements over standard ICL are largest with \textsc{Qwen3.6-27B-FP8} and generally larger on the pseudo benchmark than on the original benchmark.
Further analyses and statistical significance tests are provided in Appendix~\ref{app:stat_test}.

\section{Analysis}
\label{sec:analysis}

\begin{figure}[t]
\centering
\includegraphics[
        width=\columnwidth,
        trim=0.0cm 0.23cm 0.0cm 0.0cm,
        clip
        ]{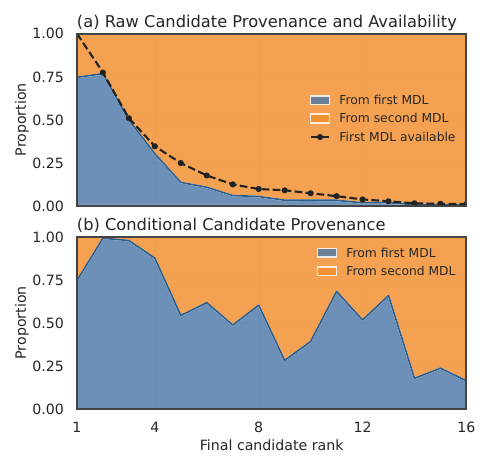}
\caption{
Provenance of the segmentation candidate list by rank (macro average over five languages).
}
\label{fig:dual-source-coverage}
\end{figure}

\begin{table}[t]
\centering
\setlength{\tabcolsep}{1.6pt}
\resizebox{\columnwidth}{!}{
\begin{tabular}{lcccccccc}
\toprule
 & \multicolumn{2}{c}{Morph.\ F1}
 & \multicolumn{2}{c}{Bnd.\ F1}
 & \multicolumn{2}{c}{Oracle@1}
 & $*$@16 
 & $*$@$K_w$\\
\cmidrule(lr){2-3}
\cmidrule(lr){4-5}
\cmidrule(lr){6-7}
System
 & macro & wtd.
 & macro & wtd.
 & macro & wtd.
 & macro & macro\\
\midrule

\multicolumn{8}{l}{\textit{(a) Standard low-resource evaluation}} \\
Dual-MDL
 & \textbf{.541} & \textbf{.500}
 & \textbf{.601} & \textbf{.591}
 & \textbf{.502} & \textbf{.436}
 & \textbf{.946} & .734\\
1st MDL only
 & .477 & .395
 & .561 & .509
 & .410 & .296
 & .747 & \textbf{.747}\\
2nd MDL only
 & .522 & .485
 & .597 & .587
 & .464 & .404
 & .921 & .724\\

\midrule
\multicolumn{8}{l}{\textit{(b) Extremely low-resource evaluation}} \\
Dual-MDL
 & \textbf{.484} & --
 & .282 & --
 & \textbf{.564} & --
 & \textbf{.950} & .696\\
1st MDL only
 & .461 & --
 & \textbf{.400} & --
 & .496 & --
 & .714 & \textbf{.714}\\
2nd MDL only
 & .464 & --
 & .295 & --
 & .529 & --
 & .945 & .661\\

\bottomrule
\end{tabular}
}
\caption{
Ablation of the two MDL runs under standard and extremely low-resource
settings. 
}
\label{tab:mdl-ablation}
\end{table}

\paragraph{Candidate List Provenance.}
Figure~\ref{fig:dual-source-coverage}(a) shows the source composition of the final candidate list, together with the proportion of words for which the first MDL provides a candidate at each rank. The first MDL accounts for 74.6\% of the final rank-1 selections and a similar proportion at rank 2. As first-MDL candidate lists are exhausted at lower ranks, the contribution of the second MDL increases. Figure~\ref{fig:dual-source-coverage}(b) considers only words for which both MDLs provide a candidate at rank $k$. Even when both candidates are available, the first MDL remains dominant at higher ranks, whereas the second MDL contributes an increasing share at lower ranks, particularly ranks 14-16.

\paragraph{Ablation of the MDL objectives.}
Table~\ref{tab:mdl-ablation} evaluates each MDL independently. 
As list lengths differ across rankers, Oracle@16 conflates candidate quality with candidate availability. We therefore report Oracle@$K_w$, where $K_w=\min\left(16,|C_1(w)|,|C_2(w)|\right)$ and $C_1(w)$ and $C_2(w)$ denote the first- and second-MDL candidate lists; all three configurations are evaluated on their top $K_w$ candidates.
In the standard setting, 
the second MDL is the stronger standalone ranker in top-1 performance and Oracle@16, while Dual-MDL performs best on all standard metrics. At matched depth, however, the first MDL attains the highest Oracle@$K_w$ (.747), suggesting that its lower Oracle@16 primarily reflects shorter candidate lists rather than lower candidate quality.
In the extremely low-resource setting, the first MDL attains the highest Boundary F1 (.400) but lower Morpheme F1 and Oracle@1. The second MDL retains substantially greater candidate coverage, while Dual-MDL again achieves the best Oracle@1 (.564) and Oracle@16 (.950). At matched depth, the first MDL again leads on Oracle@$K_w$.

\section{Conclusion}
\label{sec:conclusion}

We presented DUET, a deterministic unsupervised morphological segmenter for extreme data scarcity that separates entropy-trie candidate generation from conservative dual-MDL ranking. 
On the SIGMORPHON 2023 interlinear glossing shared task, DUET outperforms unsupervised baselines in Morpheme F1, Oracle@1, and candidate coverage under both standard and simulated extremely low-resource settings. 
Incorporating DUET segmentations into prompts also yields the best overall results among the evaluated segmentation methods on a downstream linguistic reasoning benchmark.
These results support separating structural hypothesis generation from statistical selection when corpus evidence is scarce.

We believe that DUET provides a practical and reproducible foundation for future research on low-resource languages, 
even under extreme data scarcity, 
and hope that it will prove valuable across a broader range of downstream tasks.

\section*{Limitations}
\label{sec:limitations}

Our intrinsic evaluation covers surface-recoverable gold segmentations; canonical analyses involving morphophonological alternations not recoverable from surface forms fall outside the scope of this work. 
We report a single run per condition: DUET is deterministic, while for Morfessor we fix the shuffle seed for reproducibility without estimating variability across seeds. 
Because DUET estimates branching entropy over character/atom sequences, it assumes reasonably consistent orthographic representations and identifiable word boundaries; its applicability to highly variable orthographies or scripts without explicit word segmentation remains to be evaluated. 
Finally, our downstream evaluation focuses on linguistic puzzle reasoning, leaving the utility of DUET segmentations for other tasks as future work. 
Adding segmentation candidates also increases prompt length, and the resulting gains vary across backbone models.

\section*{Ethical Statement}
\label{sec:ethical}

For all baselines, we use the official implementations and follow the settings recommended by their authors: 
Morfessor 2.0\footnote{
\url{https://github.com/aalto-speech/morfessor}
}, 
Linguistica\footnote{
\url{https://github.com/linguistica-uchicago/lxa5}
}, 
MorphAGram\footnote{
\url{https://github.com/rnd2110/MorphAGram}
},
and BPE\footnote{
\url{https://github.com/rsennrich/subword-nmt}
}.
For BPE, we additionally impose an MDL-based stopping criterion, so our implementation differs from the original release.
Additional implementation details and baseline configurations are provided in Appendix~\ref{app:implementation}.
Our main experiments use the \textit{Surface} Segmentation subset of the SIGMORPHON 2023 interlinear glossing shared-task data\footnote{
\url{https://github.com/sigmorphon/2023glossingST}
}.
For the downstream evaluation, we manually correct several annotation and formatting issues in the PuzzLing Machines Benchmark\footnote{
\url{https://ukplab.github.io/PuzzLing-Machines/}
}
and use its official evaluation script and metrics\footnote{
\url{https://ukplab.github.io/PuzzLing-Machines/data/evaluation_script.zip}
}.
Processing details are provided in Appendix~\ref{app:puzzle_data}.
All LLMs used in this work are open-weight models and are run locally.
Generation settings are provided in Appendix~\ref{app:puzzle_settings} and \ref{app:llmprompt}.

\section*{Acknowledgements}

We are grateful to all the anonymous reviewers and the area chair for their constructive feedback.
We also sincerely thank Adam Nohejl, Koki Ryu, Ryoma Kumon,
Taisei Yamamoto, and Zhihan Cao 
for their valuable discussions, suggestions, and insightful comments.

This work was supported by JST CREST Grant Number JPMJCR2565, JST BOOST Program Grant Number JPMJBY24H5, and JST SPRING Grant Number JPMJSP2108, Japan.

\bibliography{references}

\appendix
\section{The Complete Worked Example}
\label{app:example}

This section walks through every stage of \S\ref{sec:method} with concrete numbers on the running example from the main paper (Figure~\ref{fig:pipeline}): $\Vocab = \{p\,t\,s : p\in\{$\texttt{ni,u,a}$\},\ t\in\{$\texttt{na,li}$\},\ s\in\{$\texttt{soma,pika}$\}\}\cup\{$\texttt{sana}, \texttt{nini}$\}$, $N=14$, alphabet $\mathcal A=\{$\texttt{a,i,k,l,m,n,o,p,s,u}$\}$, $|\mathcal A|=10$.
The twelve grid words are real Swahili verb forms (subject prefix, tense marker, stem, final vowel); \texttt{sana} (``very'') and \texttt{nini} (``what'') are real monomorphemic words.
All numbers below are actual outputs of the algorithm.

\paragraph{Branching entropy.} Table~\ref{tab:entropy} gives the forward/backward branching entropy at every position of \texttt{ninasoma}.
At $x=$\texttt{ni}, five words continue and the observed next-atom set is $\{$\texttt{l},\texttt{n}$\}$; substituting into Eq.~\ref{eq:H} gives $H(\texttt{ni})\approx1.30$ bits,
up from $H(\texttt{n})\approx0.41$.
The backward trie contributes the sharper spikes: $h_b^2\approx1.90$ bits because, 
reading \texttt{nasoma} leftward, the preceding atom can be \texttt{i}, \texttt{u}, or \texttt{a} (the subject-prefix slot seen from the right), and $h_b^7\approx1.54$ bits
because the final \texttt{-a} can be preceded by \texttt{m}, \texttt{k}, or \texttt{n} (the stem slot before the final vowel).
That rise-then-fall on either side is exactly the boundary signal.

\begin{table}[t]\centering\small
\begin{tabular}{rlrr}
\toprule
$p$ & prefix $w_{[:p]}$ & $h_f^p$ & $h_b^p$ \\
\midrule
0 & $\varepsilon$      & 2.018 & 0.811 \\
1 & \texttt{n}         & 0.414 & 0.811 \\
2 & \texttt{ni}        & 1.296 & 1.896 \\
3 & \texttt{nin}       & 1.352 & 0.544 \\
4 & \texttt{nina}      & 1.449 & 1.287 \\
5 & \texttt{ninas}     & 0.811 & 0.371 \\
6 & \texttt{ninaso}    & 0.811 & 0.371 \\
7 & \texttt{ninasom}   & 0.811 & 1.541 \\
8 & \texttt{ninasoma}  & 0.811 & 0.660 \\
\bottomrule
\end{tabular}
\caption{Forward/backward branching entropy (bit) at every position of $w=$\texttt{ninasoma}.}
\label{tab:entropy}
\vspace{-9pt}
\end{table}

\paragraph{Candidate cut points.} Taking adjacent differences from Table~\ref{tab:entropy} into Eq.~\ref{eq:cuts}: $\Delta(2)=\max(0.882,1.352,0)=1.352$, $\Delta(4)=\max(0.097,0.916,0)=0.916$, and $\Delta(7)=\max(0,0.882,0)=0.882$ are positive local maxima, while $\Delta(3)=0.056$ is positive but shadowed by its neighbor, 
so $\mathrm{cuts}(\texttt{ninasoma})=\{2,4,7\}$: exactly the three morpheme boundaries of \texttt{ni-na-som-a}.
The same computation gives $\mathrm{cuts}(\texttt{nini})=\{2\}$ and $\mathrm{cuts}(\texttt{sana})=\{3\}$.

\paragraph{Paradigm.} Building the bipartite graph over all 14 words and running the biclique search returns 35 word-level paradigms (most small or nested inside larger ones).
The dominant one, $P_0$, pairs the circumfix-like frames $\{($\texttt{ni},\texttt{a}$), ($\texttt{u},\texttt{a}$), ($\texttt{a},\texttt{a}$)\}$ with the stems $\{$\texttt{lisom}, \texttt{lipik}, \texttt{nasom}, \texttt{napik}$\}$, a $3\times4$ biclique (Figure~\ref{fig:pipeline}).
Substituting $|\mathcal F|=3$, $|\mathcal S|=4$, $n_w=12$ into Eq.~\ref{eq:paradigmMDL} gives $C_{\text{base}}\approx338.0$, $C_{\text{model}}\approx116.3$, $C_{\text{data}}\approx43.0$, so $\Delta C\approx178.7$ bits,
far past the 1-bit bar.
In plain terms: spelling out the twelve grid words letter by letter costs about 338 bits,
while ``two small tables plus 12 (frame, stem) indices'' costs only about 159.

\paragraph{Recursive affix unpacking.} $P_0$ can take \texttt{ninasoma} only to \texttt{ni\,|\,nasom\,|\,a}; the block \texttt{nasom} is still two morphemes.
Re-running the same biclique machinery on the accepted multi-atom blocks (with their projected cut points) finds the level-1 paradigm $P_1$: the tense markers $\{$\texttt{li}, \texttt{na}$\}$ against the contexts $\{$\texttt{-som}, \texttt{-soma}, \texttt{-pik}, \texttt{-pika}, \texttt{ni-}$\}$, with $C_{\text{base}}\approx216.7$, $C_{\text{model}}\approx97.2$, $C_{\text{data}}\approx33.2$ over the recursion round's 21-unit inventory, 
so $\Delta C\approx86.3$ bits.
$P_1$ unpacks \texttt{nasom} into \texttt{na\,|\,som}, licensing the four-morpheme candidate \texttt{ni\,|\,na\,|\,som\,|\,a}; recursion then stops, since no accepted block of length $\ge2$ yields further structure.

\paragraph{Candidate tables.} After the MDL-guided search, each word's structural candidates are scored leave-one-out and retained in score order.
Table~\ref{tab:toy-cands} shows the resulting tables for the three words highlighted in Figure~\ref{fig:pipeline}.

\begin{table}[t]\centering\small
\begin{tabular}{lrl}
\toprule
Candidate & Score & Source \\
\midrule
\texttt{ni\,|\,na\,|\,som\,|\,a} & 21.87 & paradigm + recursion \\
\texttt{nina\,|\,som\,|\,a} & 8.04 & paradigm-licensed \\
\texttt{ni\,|\,na\,|\,soma} & 7.02 & paradigm-licensed \\
\texttt{ni\,|\,nasom\,|\,a} & 5.16 & paradigm $P_0$ \\
\texttt{ninasoma} & 0.00 & baseline \\
\midrule
\texttt{ni\,|\,ni} & 12.73 & paradigm-licensed \\
\texttt{nini} & 0.00 & baseline \\
\midrule
\texttt{sana} & 0.00 & baseline \\
\bottomrule
\end{tabular}
\caption{Candidate tables for \texttt{ninasoma}, \texttt{nini}, and \texttt{sana}.
Score is bits saved relative to the baseline; every non-baseline row is licensed by a discovered paradigm.}
\label{tab:toy-cands}
\vspace{-9pt}
\end{table}

Three behaviors are visible at once.
For \texttt{ninasoma}, candidates from both paradigm layers enter the same pool, and Eq.~\ref{eq:corpusMDL} ranks the fully-split analysis first: the prefix, tense, and final-vowel boundaries are each independently reused across the vocabulary, so all three pieces of evidence together are cheaper than any subset.
For \texttt{nini}, the base ranker carves \texttt{ni\,|\,ni} for $+12.73$ bits by reusing the pervasive subject marker \texttt{ni-}: a spurious split of a monomorphemic word, and exactly the over-segmentation failure mode \S\ref{sec:method-rank} targets.
For \texttt{sana}, the single entropy cut licenses no paradigm split, so the first MDL keeps only the neutral baseline; the independent construction tree can still contribute its own candidates during the final dual-MDL merge.

\paragraph{Arbitration.} Because \texttt{sana}'s top-1 saves no bits and $N=14>|\mathcal A|=10$, the tree-active condition of Eq.~\ref{eq:treeactive} holds and the conservative second opinion is consulted, with median word length $\tilde n=7$ and margin $\tau=\lceil\log_2 8\rceil=3$.
For \texttt{nini} the construction tree reads \texttt{nini} whole; the base top-1 \texttt{ni\,|\,ni} is only $2-1=1$ morpheme finer, short of the margin, so the tree wins and the final ranking becomes \texttt{nini} first, \texttt{ni\,|\,ni} second.
For \texttt{ninasoma} the tree also reads the word whole, but \texttt{ni\,|\,na\,|\,som\,|\,a} is $4-1=3$ morphemes finer, meeting the margin, so the paradigm-mandated analysis survives as top-1.
Recursion is neither ``always split deeper'' nor ``keep only the deepest path'': every layer's candidates enter the same pool, the corpus-level MDL objective decides how they combine, and the second opinion arbitrates only where paradigm evidence is thin.

\section{Arbitration-Rule Variants and Second-Ranker Selection}
\label{app:tau-ablation}

This section provides the controlled comparison behind the arbitration rule in Eq.~\ref{eq:secondopinion}.
It first defines every tested alternative
and then reports their results in the standard setting.
A final subsection keeps the rule fixed instead,
and replaces the ranker that supplies the second opinion.

\subsection{Arbitration-Rule Variants}
\subsubsection{Experimental Controls and Notation}

For compactness, write $1\equiv\mathrm{MDL\text{-}1}$ and
$2\equiv\mathrm{MDL\text{-}2}$.  For a word $w$, let
\begin{gather}
s_1(w)=s_{\mathrm{MDL\text{-}1}}^{r_1}(w),\qquad
s_2(w)=s_{\mathrm{MDL\text{-}2}}^{r_1}(w),\notag\\
\mathcal U(w)=C_1(w)\cup C_2(w),
\label{eq:app-union}
\end{gather}
where $C_j(w)$ and $r_j(s)$ are the candidate list and candidate rank used in Eq.~\ref{eq:rankmixture}.
We change only the rule that selects the final top-ranked candidate $s^\star(w)\in\mathcal U(w)$.
Candidate generation,
the two independently fitted MDL rankers,
and $K=16$ are held fixed.
The one exception is \S\ref{app:partner-merge},
which holds the rule fixed and replaces the second ranker.

Except for the union Meta-MDL variant defined below,
the remainder of the output list is constructed exactly as in the main system:
\begin{equation}
C_{\mathrm{final}}(w)=
\Big[s^\star(w)\Big]\mathbin{\Vert}
\operatorname{sort}_{q}
\big(\mathcal U(w)\setminus\{s^\star(w)\}\big),
\label{eq:app-common-tail}
\end{equation}
where $\Vert$ denotes concatenation.
We remove duplicates, truncate to $K$,
and reserve the last position for the whole-word baseline if necessary.
The order $\operatorname{sort}_{q}$ is decreasing in the rank-mixture mass $q(s)$ from Eq.~\ref{eq:rankmixture}.
Equal masses are broken, in order,
by membership in more source lists,
the lower best rank, the lower worst rank,
and the lexical representation of $s$.
The same order is used below whenever a rule calls for a rank-mass tie-break.

The fit uses every observed type in each of the five training vocabularies,
and evaluation uses the same surface subset as
Table~\ref{tab:train-results}.
We report an equal-language macro average
and an average weighted by the number of evaluated types.
Gold is consulted only after every $s^\star(w)$ and $C_{\mathrm{final}}(w)$ have been fixed.

\subsubsection{Count-Based Thresholds}

The current system contains a separate safeguard for a first-MDL top
candidate that does not beat the whole-word baseline.
Combining that safeguard with Eq.~\ref{eq:secondopinion},
the family of fixed count gates is defined by
\begin{align}
\delta(w)&=|s_1(w)|-|s_2(w)|,\notag\\
P_\tau(w)&=
\big[\mathrm{score}(s_1(w))\leq0\big]
\lor\big[\delta(w)\geq\tau\big].
\label{eq:app-fixed-predicate}
\end{align}
The corresponding choice is
\begin{equation}
s_\tau^\star(w)=
\begin{cases}
s_1(w), & P_\tau(w),\\
s_2(w), & \text{otherwise,}
\end{cases}
\label{eq:app-fixed-gate}
\end{equation}
where $|s|$ counts morphemes,
exactly as in Eq.~\ref{eq:secondopinion}.
We evaluate fixed $\tau\in\{0,1,2,3,4,5\}$.
A smaller $\tau$ makes it easier for the finer first-MDL analysis to remain first;
a larger $\tau$ delegates more cases to the conservative second MDL.

The proposed rule is the same family with a corpus-derived value
\begin{equation}
\tau=\left\lceil\log_2(\tilde n+1)\right\rceil.
\label{eq:app-current-tau}
\end{equation}
The median word lengths are $7,6,6,5,6$ for Natugu, Gitksan, Lezgi,
Nyangbo, and Uspanteko, respectively,
so Eq.~\ref{eq:app-current-tau} gives $\tau=3$ in all five full-resource vocabularies.

\subsubsection{Non-Count Arbitration Rules}

\paragraph{Rank mass.}
The first alternative removes the count gate
and directly promotes the candidate with the largest mass already used to merge the tail:
\begin{equation}
s_{\mathrm{RM}}^\star(w)
=\operatorname*{arg\,max}_{s\in\mathcal U(w)}q(s).
\label{eq:app-rank-mass}
\end{equation}
Thus a candidate is favored when it occurs near the top of either list,
and receives mass from both terms of Eq.~\ref{eq:rankmixture} when it is shared.
Ties follow the rank-mass order defined below Eq.~\ref{eq:app-common-tail}.

\paragraph{Shared candidate first.}
Let $\mathcal S(w)=C_1(w)\cap C_2(w)$.
This rule requires agreement on the exact morpheme sequence and chooses the shared candidate with the smallest sum of ranks.
Define
\begin{align}
R_{\Sigma}(s)&=r_1(s)+r_2(s),\notag\\
R_{\max}(s)&=\max\{r_1(s),r_2(s)\}.
\label{eq:app-shared-ranks}
\end{align}
Then
\begin{equation}
s_{\mathrm{SH}}^\star(w)
=\operatorname*{arg\,min}_{s\in\mathcal S(w)}
\left(R_{\Sigma}(s),R_{\max}(s)\right).
\label{eq:app-shared}
\end{equation}
The ordered pair is minimized lexicographically,
followed by the lexical representation of $s$.
If $\mathcal S(w)=\varnothing$,
we fall back to Eq.~\ref{eq:app-rank-mass}.
Because both rankers normally retain the whole-word baseline,
this rule often finds agreement by selecting no cut at all.

\paragraph{Boundary consensus.}
Exact candidate agreement may be too strict when two candidates share most but not all boundaries.
Reusing $\mathrm{cut}(s)$ from Eq.~\ref{eq:corpusMDL},
let $B_s=\mathrm{cut}(s)$ and define
\begin{equation}
J(s,t)=
\begin{cases}
\dfrac{|B_s\cap B_t|}{|B_s\cup B_t|},
& B_s\cup B_t\neq\varnothing,\\[2mm]
1, & \text{otherwise.}
\end{cases}
\label{eq:app-boundary-jaccard}
\end{equation}
Also abbreviate the reciprocal-rank weight as
\begin{equation}
\rho_j(t)=\frac{1}{r_j(t)(r_j(t)+1)}.
\label{eq:app-reciprocal-rank}
\end{equation}
The agreement score for $s\in\mathcal U(w)$ is
\begin{align}
A_{\mathrm{BC}}(s)
&=\mathbf 1[s\in C_1(w)]
\sum_{t\in C_2(w)}
J(s,t)\rho_2(t)\notag\\
&\quad+\mathbf 1[s\in C_2(w)]
\sum_{t\in C_1(w)}
J(s,t)\rho_1(t),
\label{eq:app-boundary-consensus}
\end{align}
and $s_{\mathrm{BC}}^\star(w)=\arg\max_{s\in\mathcal U(w)}
A_{\mathrm{BC}}(s)$.
The reciprocal-rank weights emphasize agreement with the other ranker's highest candidates.
Ties are broken by rank mass.

\paragraph{Normalized confidence gap.}
The two MDLs do not expose comparable absolute values:
the first uses the bit savings $\mathrm{score}(s)$ from Eq.~\ref{eq:score},
whereas the second orders candidates by a within-model construction cost $\ell_2(s)$ in natural-log
units.
We therefore compare only normalized gaps within each list.
If $C_{j,r}(w)$ is the candidate at rank $r$, define
\begin{align}
x_{1r}&=\mathrm{score}(C_{1,r}(w)),
&g_1&=x_{11}-x_{12},\notag\\
x_{2r}&=\ell_2(C_{2,r}(w)),
&g_2&=x_{22}-x_{21},
\label{eq:app-raw-gaps}
\end{align}
where the signs reflect that a larger first-MDL score but a smaller
second-MDL cost is better.
For each list, let
\begin{align}
\bar x_j&=\operatorname{median}_r x_{jr},\notag\\
\sigma_j&=\operatorname{median}_r|x_{jr}-\bar x_j|,
&D_j&=\frac{g_j}{\sigma_j}.
\label{eq:app-normalized-gap}
\end{align}
When the median absolute deviation $\sigma_j$ is zero,
its within-list range is used instead;
if that range is also zero, $D_j=0$.
A singleton list is assigned $D_j=+\infty$.  The decision is
\begin{equation}
s_{\mathrm{CG}}^\star(w)=
\begin{cases}
s_1(w), & D_1>D_2,\\
s_2(w), & D_2>D_1,\\
s_{\mathrm{RM}}^\star(w), & D_1=D_2.
\end{cases}
\label{eq:app-confidence-choice}
\end{equation}
Consequently, this rule never directly compares bits with natural-log costs.

\paragraph{Union Meta-MDL.}
Rather than choose between two top candidates,
this variant treats every member of $\mathcal U(w)$ as a candidate for a new common optimization.
It solves the same self-consistent corpus problem as the first MDL,
\begin{equation}
\widehat{\mathbf s}
=\operatorname*{arg\,min}_{
 (s_w)_{w\in\Vocab}\in\prod_{w\in\Vocab}\mathcal U(w)}
\sum_{w\in\Vocab}
\mathrm{cost}(w\leftarrow s_w),
\label{eq:app-union-meta-mdl}
\end{equation}
using Eq.~\ref{eq:corpusMDL}, its morpheme-use counts, its boundary evidence,
and the same maximal-segmentation initialization and deterministic MDL-guided search.
After the search, every $s\in\mathcal U(w)$ is scored leave-one-out
with Eq.~\ref{eq:score}, sorted, truncated to $K$,
and guaranteed a whole-word slot.
This is the only variant that reranks the entire union
instead of using the common tail in Eq.~\ref{eq:app-common-tail}.
\begin{table*}[t]
\centering
\setlength{\tabcolsep}{2.7pt}
\begin{tabular}{lcccccccc}
\toprule
 & \multicolumn{2}{c}{Morph. F1}
 & \multicolumn{2}{c}{Bnd. F1}
 & \multicolumn{2}{c}{Oracle@1}
 & \multicolumn{2}{c}{Oracle@16} \\
\cmidrule(lr){2-3}\cmidrule(lr){4-5}\cmidrule(lr){6-7}\cmidrule(lr){8-9}
Arbitration rule
 & macro & wtd. & macro & wtd. & macro & wtd. & macro & wtd. \\
\midrule
\multicolumn{9}{l}{\textit{Count-based gates}} \\
\textbf{Eq.~\ref{eq:secondopinion} (current)}
 & \textbf{.541} & \textbf{.500} & \textbf{.601} & \textbf{.591}
 & .502 & .436 & .946 & .923 \\
Fixed $\tau=0$ & .479 & .404 & .567 & .521 & .406 & .299 & .946 & .923 \\
Fixed $\tau=1$ & .489 & .436 & .580 & .554 & .415 & .330 & .946 & .923 \\
Fixed $\tau=2$ & .534 & .489 & .598 & .581 & .489 & .414 & .946 & .923 \\
Fixed $\tau=3$ & \textbf{.541} & \textbf{.500} & \textbf{.601} & \textbf{.591}
 & .502 & .436 & .946 & .923 \\
Fixed $\tau=4$ & .540 & .500 & .600 & .590 & .502 & .437 & .946 & .923 \\
Fixed $\tau=5$ & .540 & .499 & .599 & .589 & .502 & .437 & .946 & .923 \\
\midrule
\multicolumn{9}{l}{\textit{Non-count arbitration}} \\
Rank mass & .534 & .474 & .591 & .564 & .489 & .402 & .946 & .923 \\
Shared candidate first & .526 & .470 & .507 & .479 & \textbf{.533} & \textbf{.467} & .946 & .923 \\
Boundary consensus & .529 & .459 & .585 & .555 & .483 & .384 & .946 & .923 \\
Normalized confidence gap & .506 & .418 & .567 & .517 & .458 & .340 & .946 & .923 \\
Union Meta-MDL & .356 & .347 & .470 & .478 & .229 & .199 & .944 & .917 \\
Delete-one-fold stability & .535 & .458 & .592 & .554 & .485 & .379 & .946 & .923 \\
\bottomrule
\end{tabular}
\caption{Arbitration ablation on the five training
vocabularies.
\textbf{Bold} marks the strongest Morph./Bnd. F1 operating points
and the strongest Oracle@1 operating point;
Oracle@16 is primarily determined by the unchanged candidate union.}
\label{tab:tau-ablation}
\vspace{-6pt}
\end{table*}

\begin{figure*}[t]
    \centering
    \includegraphics[width=\textwidth]{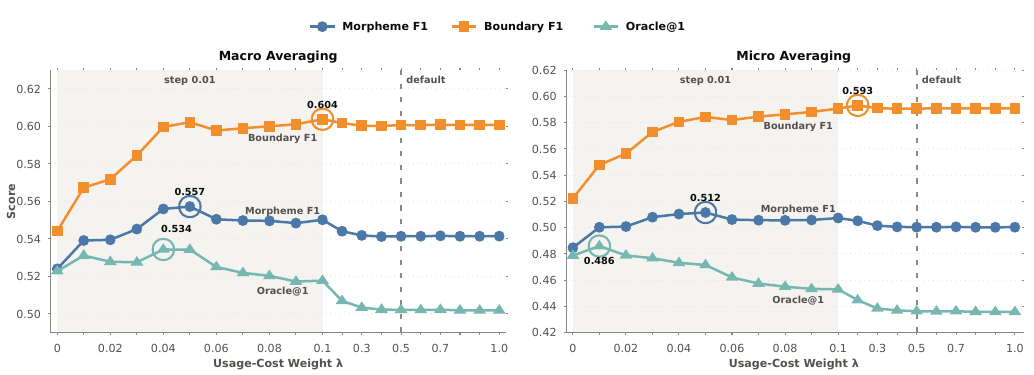}
    \caption{
        Effects of the usage-cost weight $\lambda$ on
        Morpheme F1, Boundary F1, and Oracle@1
        under macro and micro averaging.
    }
    \label{fig:lambda_full_train}
\end{figure*}

\paragraph{Delete-one-fold stability.}
Because DUET is transductive over the observed type vocabulary,
an ordinary bootstrap sample can omit a target word and leave the first MDL unable to
score it.
We instead deterministically shuffle the sorted vocabulary with seed $0$,
divide it round-robin into five blocks $F_1,\ldots,F_5$,
and refit both MDLs on each $\Vocab\setminus F_f$.
Let $s_{j,-f}^{r_1}(w)$ be ranker $j$'s top candidate in that refit.
The exact candidate vote is
\begin{equation}
V_w(s)=
\sum_{f:w\notin F_f}\ \sum_{j\in\{1,2\}}
\mathbf 1\!\left[s=s_{j,-f}^{r_1}(w)\right].
\label{eq:app-stability-vote}
\end{equation}
Each word remains in four refits and therefore receives eight votes.  We
choose
\begin{equation}
s_{\mathrm{ST}}^\star(w)
=\operatorname*{arg\,max}_{s\in\mathcal U(w)}V_w(s),
\label{eq:app-stability-choice}
\end{equation}
with rank mass breaking vote ties.  Restricting the choice to
$\mathcal U(w)$ ensures that the selected result can enter the unchanged
full-data candidate list.

\subsubsection{Results}

Table~\ref{tab:tau-ablation} shows a clear knee at $\tau=3$.
With $\tau=0$ or $1$,
Eq.~\ref{eq:app-fixed-gate} preserves the finer first-MDL top candidate too readily,
reproducing its over-segmentation failure mode and reducing every top-1 metric.
Fixed $\tau=2$ recovers most of the loss.
Increasing the gate to $4$ or $5$ delegates slightly more cases to the conservative second MDL but changes little.
The proposed rule and fixed $\tau=3$ are identical here
and give the strongest overall Morpheme/Boundary F1 operating point.

Rank mass and stability are the closest non-count alternatives in macro F1,
but their weighted results are lower,
showing that either cross-list rank agreement or resampling stability alone is not a reliable proxy for the desired segmentation.
Boundary consensus also remains below the current rule on every top-1 metric.
The normalized confidence gap performs worse still:
normalization makes the two within-model margins dimensionless,
but does not make their different score geometries equally calibrated.

The shared-candidate rule illustrates an important competing operating point.
It raises Oracle@1 from $.502/.436$ to $.533/.467$,
but selects a whole-word analysis for $56.5\%$ of evaluated types,
compared with $28.8\%$ under Eq.~\ref{eq:secondopinion}.
Its macro boundary recall therefore falls from $.614$ to $.410$,
and its Boundary F1 falls from $.601/.591$ to $.507/.479$.
The exact-match gain is consequently driven
by a conservative whole-word bias rather than a uniform improvement in segmentation.

\begin{table*}[t]
\centering
\resizebox{\textwidth}{!}{%
\begin{tabular}{l|rrrrrrr|rrrrrrr|rrrrrrr}
\toprule

& \multicolumn{7}{c|}{\textbf{Morpheme F1}}
& \multicolumn{7}{c|}{\textbf{Boundary F1}}
& \multicolumn{7}{c}{\textbf{Oracle@1}} \\

\cmidrule(lr){2-8}
\cmidrule(lr){9-15}
\cmidrule(lr){16-22}

System
& Git & Lez & Nat & Nya & Usp & Macro & Wtd
& Git & Lez & Nat & Nya & Usp & Macro & Wtd
& Git & Lez & Nat & Nya & Usp & Macro & Wtd \\

\midrule

whole-word
& .337 & .573 & .389 & .213 & .409 & .384 & .369
& .000 & .000 & .000 & .000 & .000 & .000 & .000
& .453 & .697 & .567 & .312 & .556 & .517 & .503 \\

every-atom
& .093 & .039 & .029 & .129 & .122 & .082 & .082
& .276 & .175 & .314 & .399 & .304 & .294 & .309
& .009 & .010 & .007 & .010 & .011 & .010 & .010 \\

\midrule

BPE
& .183 & .094 & .346 & .346 & .137 & .221 & .201
& .338 & .172 & .487 & .509 & .304 & .362 & .353
& .009 & .020 & .164 & .179 & .023 & .079 & .070 \\

Linguistica
& .336 & .514 & .375 & .303 & .448 & .395 & .388
& .095 & .068 & .228 & .233 & .233 & .171 & .190
& .462 & \textbf{.646} & .542 & .351 & \textbf{.560} & .512 & .500 \\

MorphAGram
& .553 & .235 & .406 & \textbf{.547} & .279 & .404 & .406
& \textbf{.637} & .270 & \textbf{.541} & \textbf{.645} & .402 & \textbf{.499} & \textbf{.505}
& .479 & .232 & .306 & .466 & .188 & .334 & .334 \\

Morfessor
& .606 & .501 & .381 & .543 & .423 & .491 & .478
& .605 & .403 & .295 & .604 & \textbf{.443} & .470 & .459
& .615 & .566 & .453 & \textbf{.489} & .447 & .514 & .496 \\

DUET
& \textbf{.624} & \textbf{.562} & \textbf{.513} & .407 & \textbf{.463} & \textbf{.514} & \textbf{.489}
& .609 & \textbf{.438} & .479 & .410 & .424 & .472 & .450
& \textbf{.632} & .633 & \textbf{.550} & .410 & .523 & \textbf{.550} & \textbf{.524} \\

\bottomrule
\end{tabular}
}
\caption{
Language-wise results on the development set.
}
\label{tab:dev-all}
\end{table*}

\begin{table*}[t]
\centering
\resizebox{\textwidth}{!}{%
\begin{tabular}{l|rrrrrrr|rrrrrrr|rrrrrrr}
\toprule

& \multicolumn{7}{c|}{\textbf{Morpheme F1}}
& \multicolumn{7}{c|}{\textbf{Boundary F1}}
& \multicolumn{7}{c}{\textbf{Oracle@1}} \\

\cmidrule(lr){2-8}
\cmidrule(lr){9-15}
\cmidrule(lr){16-22}

System
& Git & Lez & Nat & Nya & Usp & Macro & Wtd
& Git & Lez & Nat & Nya & Usp & Macro & Wtd
& Git & Lez & Nat & Nya & Usp & Macro & Wtd \\

\midrule

whole-word
& .312 & .524 & .385 & .226 & .311 & .352 & .339
& .000 & .000 & .000 & .000 & .000 & .000 & .000
& .455 & .650 & .561 & .330 & .440 & .487 & .472 \\

every-atom
& .027 & .059 & .027 & .139 & .148 & .080 & .094
& .269 & .207 & .320 & .402 & .330 & .306 & .322
& .000 & .011 & .008 & .011 & .009 & .008 & .009 \\

\midrule

BPE
& .160 & .126 & .390 & .385 & .213 & .255 & .260
& .332 & .218 & .524 & .534 & .383 & .398 & .412
& .014 & .058 & .198 & .231 & .055 & .111 & .106 \\

Linguistica
& .374 & .476 & .373 & .321 & .399 & .389 & .383
& .350 & .024 & .197 & .241 & .273 & .217 & .214
& .448 & \textbf{.617} & \textbf{.540} & .376 & \textbf{.484} & .493 & .486 \\

MorphAGram
& .370 & .293 & .444 & .376 & .404 & .377 & .385
& .519 & .359 & .563 & .554 & \textbf{.529} & .505 & .515
& .252 & .259 & .357 & .215 & .311 & .279 & .282 \\

Morfessor
& .599 & .472 & .484 & .497 & .443 & .499 & .484
& \textbf{.669} & .441 & .560 & \textbf{.574} & .515 & \textbf{.552} & \textbf{.542}
& .538 & .507 & .410 & .435 & .405 & .459 & .442 \\

DUET
& \textbf{.605}
& \textbf{.519}
& \textbf{.553}
& \textbf{.501}
& \textbf{.454}
& \textbf{.526}
& \textbf{.511}
& \textbf{.669}
& \textbf{.463}
& \textbf{.577}
& .524
& .498
& .546
& .531
& \textbf{.559}
& .562
& .534
& \textbf{.490}
& .444
& \textbf{.518}
& \textbf{.505} \\

\bottomrule
\end{tabular}
}
\caption{
Language-wise results on the test set.
}
\label{tab:test-all}
\vspace{-6pt}
\end{table*}

\begin{figure*}[t]\centering
\includegraphics[width=\textwidth]{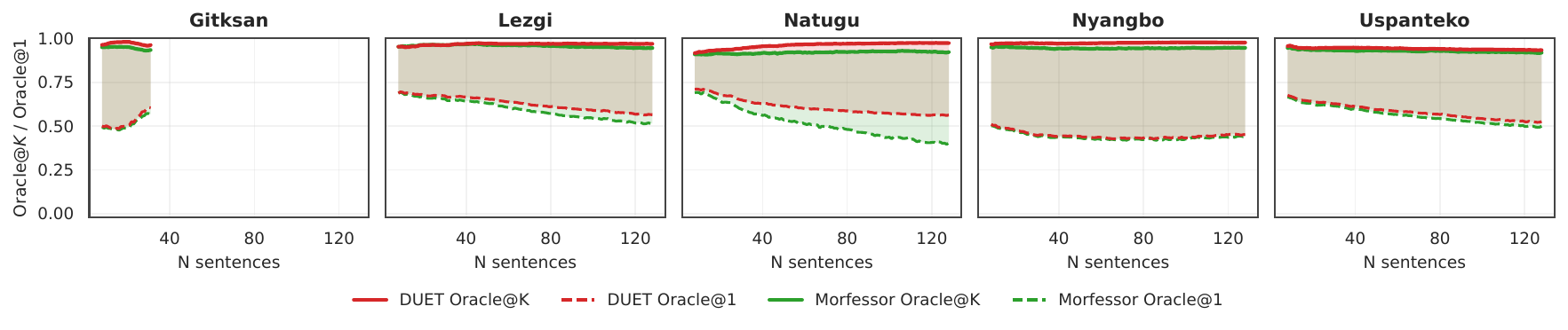}
\caption{Candidate \emph{generation} vs.\ \emph{ranking} in the extremely low-resource sweep: Oracle@$K$ (solid) vs.\ Oracle@1 (dashed) for DUET (red) and Morfessor (green); the shaded gap is the ranking margin discussed in \S\ref{app:lowres-gap}.}
\label{fig:lowres-gap}
\end{figure*}

Union Meta-MDL exhibits the opposite behavior.
Its macro boundary recall is $.800$,
but its precision is only $.340$,
yielding substantially poorer Morpheme F1, Boundary F1, and Oracle@1.
Giving the first MDL's corpus objective unrestricted access to the second MDL's additional splits
thus reintroduces over-segmentation rather than producing a better common scale.
Oracle@16 is nearly invariant for the other alternatives
because they retain the same union
and the same $q$-ranked tail apart from the promoted top candidate.
The Meta-MDL can change which members survive truncation
and is the only rule whose Oracle@16 decreases visibly.

Finally, we retain $\tau=\lceil\log_2(\tilde n+1)\rceil$
because it adapts to the word-length scale
without introducing a language-specific parameter
and yields the best overall Morpheme/Boundary F1 trade-off in this comparison.
Although it is equivalent to a fixed threshold of $\tau=3$
for all languages evaluated in our experiments,
the adaptive form captures the structural intuition that
shorter words offer fewer potential segmentation points
and therefore warrant a smaller arbitration margin,
whereas longer words permit more candidate splits
and warrant a larger one.

\subsection{Substituting the Second Opinion}
\label{app:partner-merge}

The variants above change the arbitration rule and keep both MDL rankers.
This subsection replaces the second candidate list $C_2(w)$
by the output of an existing segmenter.
Every segmenter is fitted on the same vocabulary the second MDL is fitted on.
Morfessor returns a 16-best list and therefore enters the rank mixture
on the same terms as the second MDL;
BPE, Linguistica and MorphAGram return a single candidate each.
MorphAGram is run in live sampling mode.
Table~\ref{tab:partner-merge} reports both evaluation regimes.

\begin{table}[t]
\centering
\small
\setlength{\tabcolsep}{2.7pt}
\resizebox{\columnwidth}{!}{
\begin{tabular}{lcccccccc}
\toprule
 & \multicolumn{2}{c}{Morph.\ F1}
 & \multicolumn{2}{c}{Bnd.\ F1}
 & \multicolumn{2}{c}{Oracle@1}
 & \multicolumn{2}{c}{Oracle@16} \\
\cmidrule(lr){2-3}\cmidrule(lr){4-5}\cmidrule(lr){6-7}\cmidrule(lr){8-9}
Second opinion
 & macro & wtd. & macro & wtd. & macro & wtd. & macro & wtd. \\
\midrule
\multicolumn{9}{l}{\textit{(a) Standard low-resource evaluation}} \\
None (1st MDL only)
 & .477 & .395 & .561 & .509 & .410 & .296 & .747 & .696 \\
\textbf{2nd MDL (current)}
 & .541 & .500 & .601 & .591 & .502 & .436 & .946 & .923 \\
Morfessor (16-best)
 & .540 & .504 & .601 & .595 & .500 & .439 & .933 & .912 \\
MorphAGram (1-best)
 & .498 & .497 & .583 & .599 & .449 & .425 & .803 & .767 \\
BPE (1-best)
 & .370 & .363 & .469 & .482 & .308 & .260 & .787 & .743 \\
Linguistica (1-best)
 & .337 & .332 & .266 & .353 & .412 & .367 & .754 & .710 \\
\midrule
\multicolumn{9}{l}{\textit{(b) Extremely low-resource evaluation}} \\
None (1st MDL only)
 & .461 & -- & .400 & -- & .496 & -- & .714 & -- \\
\textbf{2nd MDL (current)}
 & .484 & -- & .282 & -- & .564 & -- & .950 & -- \\
Morfessor (16-best)
 & .486 & -- & .285 & -- & .564 & -- & .944 & -- \\
MorphAGram (1-best)
 & .403 & -- & .354 & -- & .455 & -- & .754 & -- \\
BPE (1-best)
 & .353 & -- & .368 & -- & .412 & -- & .730 & -- \\
Linguistica (1-best)
 & .453 & -- & .127 & -- & .577 & -- & .718 & -- \\
\bottomrule
\end{tabular}
}
\caption{
Replacing the ranker that supplies the second opinion.
The first MDL and the arbitration rule of Eq.~\ref{eq:secondopinion}
are unchanged; only $C_2(w)$ differs.
(a) uses the full training vocabulary, as in Table~\ref{tab:tau-ablation}.
(b) follows the extremely low-resource protocol of Table~\ref{tab:lowres-main},
for which weighted averages are not reported.
}
\label{tab:partner-merge}
\end{table}

\subsection{Parameter for Balancing the Two Terms in the First MDL}
\label{app:eq6_lambda}

We conduct additional experiments by introducing a weighting parameter $\lambda$ to balance the contributions of $C_{\mathrm{usage}}$ and $B_{\mathrm{bnd}}$.
Accordingly, we reformulate Eq.~\ref{eq:corpusMDL} as follows:

\begin{equation}
    \lambda \cdot C_{\mathrm{usage}}(\cdot)
    - (1-\lambda)\cdot B_{\mathrm{bnd}}(\cdot)
    \label{eq:usage_balance}
\end{equation}

The experimental results are presented in Figure~\ref{fig:lambda_full_train}.
Oracle@16 is omitted from the figure, as it remains unchanged across all values of $\lambda$.
We observe that $\lambda$ indeed affects the other three evaluation metrics. 
Specifically, smaller values of $\lambda$ generally yield better performance on Morpheme F1 and Oracle@1,
whereas Boundary F1 tends to improve as $\lambda$ increases within the range of $[0, 0.1]$. 
Although DUET is designed as a parameter-free method, 
we also provide $\lambda$ as an optional hyperparameter to allow users to adjust the balance between the two terms.
Its default value is set to $0.5$,
assigning equal weight to both terms.

\section{Implementation and Baseline Configuration}
\label{app:implementation}

The five non-LLM systems below are fitted transductively on the same normalized word list on which they are evaluated.
Table~\ref{tab:implementation} records their fixed implementation choices used for the reported results.
Quantities that depend on the input vocabulary (e.g., DUET's alphabet, frame inventory, and data-adaptive merge count) are recomputed for every word list; they are not selected on a development set.
% The evaluation and oracle bound is $K=16$ for every system that exposes an $n$-best list.

\begin{table*}[t]
\centering\small
\renewcommand{\arraystretch}{1.18}
\begin{tabular}{p{0.14\textwidth}p{0.80\textwidth}}
\toprule
System & Configuration used in all reported non-LLM experiments \\
\midrule
BPE & Same atom-level input and type count $1$ per word. Greedily merge the most frequent adjacent pair (lexicographic tie-break);
choose the retained merge count by the two-part MDL minimum,
using add-$\tfrac12$ smoothing in the corpus code.
Training stops when no pair recurs, after $10$ merges
without a new MDL minimum, or at $2000$ merges.
BPE returns one analysis. \\
\addlinespace
Linguistica & The \texttt{lxa5} \texttt{linguistica} package \texttt{5.3.0},
called as \texttt{from\_wordlist} with the library's default parameters.
We read \texttt{broken\_words\_right\_to\_left()}, which is lxa5's shared-affix
trie parse rather than its signature module, so the only parameter in effect is
\texttt{min\_stem\_length}$=4$: a word is cut wherever the reversed, sorted type
list shares a suffix of at least that length. Words the parse does not cover are
left unsegmented. Linguistica returns one analysis. \\
\addlinespace
MorphAGram & The standard Prefix--Stem--Suffix Adaptor Grammar,
using Mark Johnson's \texttt{py-cfg} source release \texttt{2013-09-23}. The sampler
uses \texttt{-P -D -E -R -1 -d 0}, $a=10^{-4}$, $b=10^4$,
$e=f=1$, $g=10$, and $h=0.1$; it runs $1000$ iterations with seed $1$.
ASLR is disabled before launching \texttt{py-cfg} to make this seeded sampler reproducible.
Full-split evaluation reads segmentations produced by this configuration;
the extremely low-resource sweep runs it live once per sampled word list.
We extract the top-level Prefix, Stem, and Suffix nodes,
yielding one analysis. \\
\addlinespace
Morfessor & Morfessor \texttt{2.0.6}; \texttt{BaselineModel} with its default
settings, type count $1$ per observed word, and \texttt{train\_batch()}.
Before every fit we set the library's global shuffle seed to $0$. We decode
with 1-best Viterbi and retain up to $16$ distinct analyses from
\texttt{viterbi\_nbest}. \\
\addlinespace
DUET & NFC normalization and casefolding (on); atom-level input;
Jeffreys smoothing $\alpha=0.5$; positive local entropy maxima as cut
candidates, including edge-adjacent positions only when supported by the
same peak criterion; retain paradigms only when $\Delta C\geq1$
bit.
We request $K=16$ candidates (the implementation also enforces
$K\geq\lceil\log_2 N\rceil$). The MDL-guided search uses a fixed lexical order.
When its conservative construction-tree second opinion is invoked, its
shuffle seed is $0$ and it stops below $0.005$ nat per corpus boundary or
after $50$ epochs. \\
\bottomrule
\end{tabular}
\caption{Implementation and baseline configuration.
``Default'' means the named library's own default, after the settings shown here have been fixed.}
\label{tab:implementation}
\vspace{-6pt}
\end{table*}

\section{Additional Segmentation Results}
\label{app:seg-results}

This section collects the segmentation results omitted from the main paper for space:
the full per-language dev and test numbers,
results on Arapaho and Tsez from the SIGMORPHON 2023 interlinear glossing shared task,
the candidate-generation/ranking decomposition of the low-resource sweep,
and an additional test on a disjoint language cohort.

\subsection{Dev / Test Per-Language Results}
\label{app:devtest}

\begin{figure*}[t]\centering
\includegraphics[width=\textwidth]{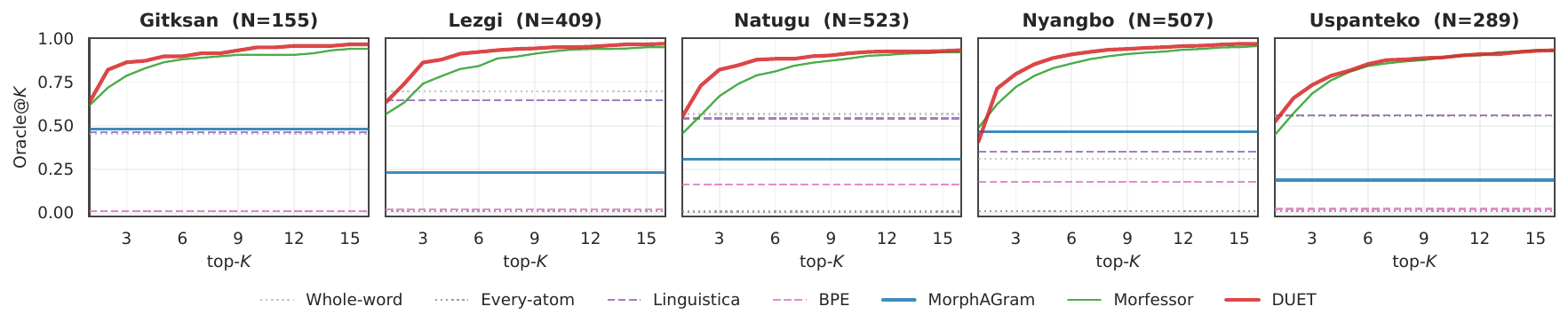}
\caption{Development set Oracle@$K$ vs.\ $K$ (same format as Figure~\ref{fig:oracle-train} in the main content).}
\label{fig:oracle-dev}
\end{figure*}

\begin{figure*}[t]\centering
\includegraphics[width=\textwidth]{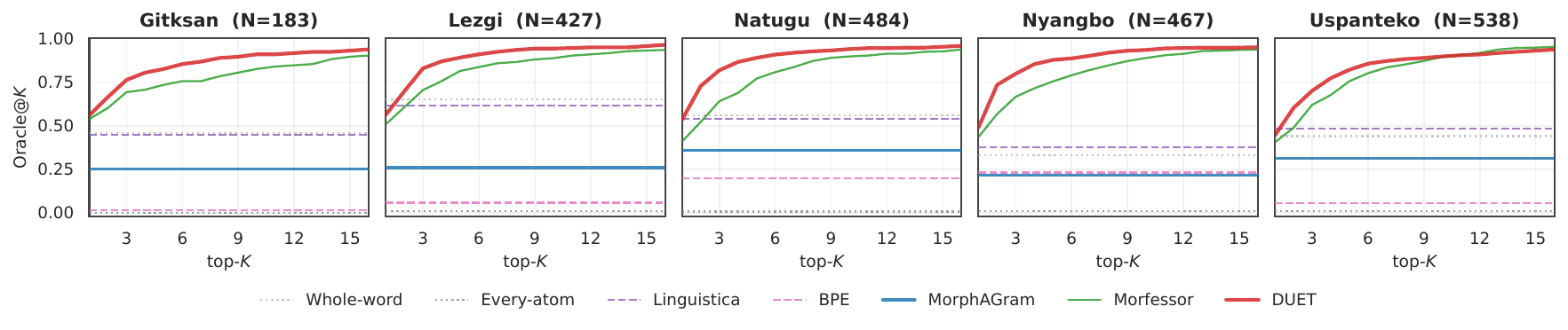}
\caption{Test set Oracle@$K$ vs.\ $K$ (same format as Figure~\ref{fig:oracle-train} in the main content).}
\label{fig:oracle-test}
\end{figure*}

Tables~\ref{tab:dev-all}, \ref{tab:test-all}
and Figures~\ref{fig:oracle-dev}, \ref{fig:oracle-test} give the full per-language results on the dev and test splits, in the same format as the train-split table and oracle curve in the main paper.

\subsection{Results on Arapaho and Tsez}
\label{app:ara_tsez}

Arapaho and Tsez, the two languages with the largest datasets in the SIGMORPHON 2023 interlinear glossing shared task, 
are selected for the development and exploration of our DUET method. 
Consequently, their results are reported separately from those of the other five languages.
We use the training, development, 
and test sets provided for both languages, 
yielding a total of six datasets.
Due to their large sizes, 
we perform sentence-level random sampling on the training,
development, and test sets of Arapaho and the training set of Tsez,
using 64 random seeds and targeting approximately 1,200 word types per sampled subset. 
Detailed statistics for the full datasets and the sampled subsets are presented in Table~\ref{tab:ara_tsez}.

\begin{table}[htbp]
\centering
\resizebox{\columnwidth}{!}{
\begin{tabular}{lcrrrr}
\toprule
Language & Type & Sentences & Word types & Surface & Cover. \\
\midrule
\multicolumn{6}{l}{\textit{(a) Full datasets}} \\
\multirow{3}{*}{Arapaho}
& train & 36,123 & 41,199 & 34,407 & 84\% \\
& dev   & 4,892  & 7,630  & 6,551  & 86\% \\
& test  & 4,892  & 7,585  & 6,551  & 86\% \\
\cdashline{1-6}
\multirow{3}{*}{Tsez}
& train & 3,547  & 8,540  & 4,329  & 51\% \\
& dev$\dagger$   & 445    & 2,328  & 1,270  & 55\% \\
& test$\dagger$  & 445    & 2,233  & 1,256  & 56\% \\
\midrule
\multicolumn{6}{l}{\textit{(b) Sampled subsets}} \\
\multirow{3}{*}{Arapaho}
& train$\dagger$ & 534 & 1,229 & 1,090 & 89\% \\
& dev$\dagger$   & 557 & 1,205 & 1,067 & 89\% \\
& test$\dagger$  & 557 & 1,207 & 1,072 & 89\% \\
\cdashline{1-6}
Tsez
& train$\dagger$ & 190  & 1,200  & 707  & 59\% \\
\bottomrule
\end{tabular}
}
\caption{Dataset statistics for Arapaho and Tsez from the SIGMORPHON 2023 shared task. 
Statistics for the sampled subsets are averaged over 64 random seeds.
$\dagger$ denotes the datasets used in our experiments.}
\label{tab:ara_tsez}
\end{table}

As shown in Table~\ref{tab:ara_tsez_results}, 
during the development of DUET, 
our objective was not to outperform Morfessor across all evaluation metrics; 
instead, we placed greater emphasis on Oracle@16.
We also report results aggregated across all seven languages from the SIGMORPHON 2023 shared task.
Given the imbalanced dataset sizes across languages, we consider only macro-averaged results for this comparison. 
Consistent with the results reported in Table~\ref{tab:train-results}, 
DUET still outperforms Morfessor across all evaluation metrics.

\begin{table}[htbp]
\centering
% \small
\setlength{\tabcolsep}{4pt}
\resizebox{\columnwidth}{!}{
\begin{tabular}{lcrrrr}
\toprule
Lang. & Type & Morph.\ F1 & Bnd.\ F1 & Oracle@1 & Oracle@16 \\
\midrule
\multirow{3}{*}{Arp}
& train & \textbf{.291} / .290 & .239 / \textbf{.269} & .255 / \textbf{.258} & \textbf{.558} / .411 \\
& dev   & \textbf{.274} / .255 & .226 / \textbf{.253} & .255 / \textbf{.260} & \textbf{.562} / .435 \\
& test  & \textbf{.286} / .271 & .241 / \textbf{.264} & .266 / .266 & \textbf{.566} / .427 \\
\midrule
\multirow{3}{*}{Tsz}   
& train & .531 / \textbf{.535} & \textbf{.498} / .496 & \textbf{.487} / .476 & \textbf{.906} / .898 \\
& dev   & .585 / \textbf{.600} & .539 / \textbf{.546} & \textbf{.519} / .514 & \textbf{.896} / .882 \\
& test  & \textbf{.588} / .563 & \textbf{.540} / .508 & \textbf{.513} / .474 & \textbf{.896} / .854 \\
\midrule
All  
& train & \textbf{.541} / .519 & \textbf{.601} / .597 & \textbf{.502} / .460 & \textbf{.947} / .879 \\
\bottomrule
\end{tabular}
}
\caption{
Results for Arapaho and Tsez from the SIGMORPHON 2023 shared task.
Values are reported as DUET / Morfessor, 
with the better-performing score shown in \textbf{bold}.
``All'' denotes the macro-average results considering all seven languages.}
\label{tab:ara_tsez_results}
\end{table}

\subsection{Candidate Generation vs.\ Ranking in the Low-Resource Sweep}
\label{app:lowres-gap}

Figure~\ref{fig:lowres-gap} decomposes the low-resource results of \S\ref{sec:lowres}
into the two abilities that Oracle@$K$ and Oracle@1 measure separately.
For both DUET and Morfessor the two curves stay far apart at every sampling budget:
the correct segmentation is almost always present in the candidate list,
and the shaded gap is the share of words that the ranker fails to surface as its top choice.
The remaining headroom in the low-resource regime is therefore
in ranking rather than in candidate generation.

\subsection{Results on SIGMORPHON 2022}
\label{app:blind2022}

We also conduct experiments on the SIGMORPHON 2022\footnote{\url{https://github.com/sigmorphon/2022SegmentationST}} nine-language word-level test set \citep{sigmorphon2022} 
by randomly sampling 5,000 words from datasets containing more than 10,000 words using 32 random seeds.
Notably, unlike the SIGMORPHON 2023 shared task,
the SIGMORPHON 2022 shared task does not exclusively cover inherently low-resource languages, 
and sampling is performed at the word level rather than the sentence level.

\begin{table}[htbp]
\centering
% \small
\setlength{\tabcolsep}{4pt}
\resizebox{\columnwidth}{!}{
\begin{tabular}{lcrrrr}
\toprule
Lang. & Surface & Morph.\ F1 & Bnd.\ F1 & Oracle@1 & Oracle@16 \\
\midrule
Ces & 4,000 & .217 / \textbf{.228} & .407 / \textbf{.420} & .069 / \textbf{.071} & \textbf{.446} / .418 \\
Eng & 3,517 & \textbf{.223} / .213 & \textbf{.232} / .152 & \textbf{.252} / .249 & \textbf{.844} / .805 \\
Fra & 1,022 & .166 / .166 & .192 / \textbf{.198} & \textbf{.153} / .145 & \textbf{.512} / .283 \\
Hun & 3,244 & .349 / .349 & \textbf{.529} / .527 & .183 / \textbf{.188} & \textbf{.651} / .541 \\
Ita & 388   & \textbf{.208} / .187 & \textbf{.175} / .174 & \textbf{.249} / .207 & \textbf{.735} / .291 \\
Lat & 4,910 & .083 / .083 & \textbf{.232} / .231 & \textbf{.035} / .033 & \textbf{.377} / .159 \\
Mon & 663   & \textbf{.623} / .622 & \textbf{.660} / .659 & \textbf{.591} / .590 & \textbf{.944} / .916 \\
Rus & 551   & \textbf{.137} / .135 & \textbf{.193} / .185 & \textbf{.090} / .089 & \textbf{.788} / .412 \\
Spa & 586   & \textbf{.096} / .091 & \textbf{.101} / .092 & \textbf{.103} / .096 & \textbf{.434} / .198 \\
\midrule
All & -     & \textbf{.234} / .231 & \textbf{.302} / .293 & \textbf{.191} / .186 & \textbf{.637} / .447 \\
\bottomrule
\end{tabular}
}
\caption{
Results on the sub-test set from the SIGMORPHON 2022 shared task.
Values are reported as DUET / Morfessor, 
with the better-performing score shown in \textbf{bold}.
``All'' denotes the macro-average results across all nine languages.}
\label{tab:sigmor_2022}
\end{table}

As shown in Table~\ref{tab:sigmor_2022}, 
both DUET and Morfessor achieve low scores on Morpheme F1, Boundary F1, and Oracle@1 across all languages except  Mongolian,
with particularly poor performance on Oracle@1. 
This suggests that, for inherently high-resource languages, 
sampling further exacerbates the sparsity of linguistic cues necessary for morphological segmentation, 
leading to poor performance for both methods.
Consequently, simulating low-resource conditions by sampling high-resource languages may not provide a suitable evaluation setting. 
However, we observe that DUET substantially outperforms Morfessor on Oracle@16 even under these challenging conditions.
This is because Morfessor tends to produce overly conservative segmentations when segmentation cues are insufficient, 
whereas DUET's dual-MDL strategy helps balance the two extremes
of over-segmentation and overly conservative segmentation.

\section{PuzzLing Machines: Experimental Details}
\label{app:puzzling_machines}

\subsection{Data Information}
\label{app:puzzle_data}
The PuzzLing Machines Benchmark contains only the machine translation task,
the most common category of linguistic puzzles.
Each puzzle is formatted with three components:
meta-information~(hints, which may be absent),
a context of parallel example sentence pairs~(aka., \textit{Rosetta Stone}; \citealp{bozhanov-derzhanski-2013-rosetta}),
and a test set containing translation queries in either the en$\rightarrow$xx or xx$\rightarrow$en direction.

Starting from the original benchmark,
we manually reviewed every puzzle and corrected a variety of issues,
including missing meta-information,
inconsistent Unicode symbols~(e.g., quotation marks),
incomplete sets of valid references,
and incorrect gold answers.
Of the original 86 puzzles,
81 were retained after this curation process, covering 72 languages.
The resulting benchmark contains 648 translation questions
(426 en$\rightarrow$xx and 222 xx$\rightarrow$en),
with an average of 16.3 parallel context pairs and 8 translation questions per puzzle.

\begin{table*}[t]
\centering
\resizebox{\textwidth}{!}{%
\begin{tabular}{llccc@{\hspace{5pt}}ccc@{\hspace{10pt}}cccc@{\hspace{5pt}}cccc}
\toprule

\multirow{3}{*}{Model} &
\multirow{3}{*}{Metric} &
\multicolumn{6}{c}{\textbf{Overall}} &
\multicolumn{8}{c}{\textbf{en$\rightarrow$xx}}
\\
\cmidrule(lr){3-8}
\cmidrule(lr){9-16}
& &
\multicolumn{3}{c}{DUET vs.\ ICL} &
\multicolumn{3}{c}{DUET vs.\ Morfessor} &
\multicolumn{4}{c}{DUET vs.\ ICL} &
\multicolumn{4}{c}{DUET vs.\ Morfessor}
\\
\cmidrule(lr){3-5}
\cmidrule(lr){6-8}
\cmidrule(lr){9-12}
\cmidrule(lr){13-16}
&&$\Delta$&Win&$p$&$\Delta$&Win&$p$&ICL&$\Delta$&Win&$p$&Morf.&$\Delta$&Win&$p$
\\
\midrule
\multirow{4}{*}{\textsc{Qwen3.6}}
& BLEU-2
& \textbf{+1.87} & \textbf{95.04\%} & \textbf{0.050}
& +1.25 & 87.88\% & 0.121
& 50.63 & \textbf{+3.04} & \textbf{97.98\%} & \textbf{0.020}
& 51.37 & \textbf{+2.30} & \textbf{95.58\%} & \textbf{0.044}
\\

& chrF-3
& +0.70 & 86.30\% & 0.137
& +0.63 & 83.68\% & 0.163
& 75.55 & \textbf{+1.27} & \textbf{97.10\%} & \textbf{0.029}
& 75.60 & \textbf{+1.22} & \textbf{95.48\%} & \textbf{0.045}
\\

& $1-$cTER
& +0.18 & 60.00\% & 0.400
& +0.42 & 70.22\% & 0.298
& 75.41 & +0.74 & 84.70\% & 0.153
& 75.22 & +0.92 & 85.69\% & 0.143
\\

& EM
& \textbf{+2.19} & \textbf{95.49\%} & \textbf{0.045}
& +1.79 & 91.83\% & 0.082
& 34.55 & \textbf{+3.24} & \textbf{97.34\%} & \textbf{0.027}
& 34.84 & \textbf{+2.96} & \textbf{97.37\%} & \textbf{0.026}
\\
\midrule

\multirow{4}{*}{\textsc{Gemma4}}

& BLEU-2
& \textbf{+2.64} & \textbf{97.70\%} & \textbf{0.023}
& \textbf{+2.51} & \textbf{98.59\%} & \textbf{0.014}
& 58.75 & \textbf{+3.10} & \textbf{96.71\%} & \textbf{0.033}
& 58.96 & \textbf{+2.89} & \textbf{98.06\%} & \textbf{0.019}
\\

& chrF-3
& +0.39 & 70.05\% & 0.299
& \textbf{+1.35} & \textbf{96.04\%} & \textbf{0.040}
& 81.27 & +0.09 & 54.11\% & 0.459
& 79.81 & \textbf{+1.55} & \textbf{95.54\%} & \textbf{0.045}
\\

& $1-$cTER
& +0.55 & 72.44\% & 0.276
& \textbf{+1.70} & \textbf{96.90\%} & \textbf{0.031}
& 81.57 & +0.10 & 53.82\% & 0.462
& 79.86 & \textbf{+1.82} & \textbf{95.57\%} & \textbf{0.044}
\\

& EM
& \textbf{+3.73} & \textbf{99.08\%} & \textbf{0.009}
& \textbf{+2.25} & \textbf{95.39\%} & \textbf{0.046}
& 44.23 & \textbf{+3.80} & \textbf{97.85\%} & \textbf{0.021}
& 45.26 & \textbf{+2.77} & \textbf{95.89\%} & \textbf{0.041}
\\
\midrule

\multirow{4}{*}{\textsc{GPT-oss}}

& BLEU-2
& +1.53 & 89.04\% & 0.110
& +0.35 & 60.02\% & 0.400
& 42.68 & \textbf{+3.06} & \textbf{97.76\%} & \textbf{0.022}
& 44.29 & +1.45 & 81.96\% & 0.180
\\

& chrF-3
& +0.02 & 50.65\% & 0.494
& -0.16 & 42.24\% & 0.578
& 67.99 & +0.78 & 77.79\% & 0.222
& 68.10 & +0.66 & 74.12\% & 0.259
\\

& $1-$cTER
& -0.29 & 38.56\% & 0.614
& -0.21 & 41.07\% & 0.589
& 66.31 & +0.62 & 69.25\% & 0.308
& 66.09 & +0.85 & 73.00\% & 0.270
\\

& EM
& +1.64 & 89.38\% & 0.106
& +0.56 & 64.62\% & 0.354
& 26.24 & \textbf{+3.47} & \textbf{98.96\%} & \textbf{0.010}
& 27.89 & +1.83 & 85.42\% & 0.146
\\
\bottomrule
\end{tabular}
}
\caption{
Hierarchical bootstrap test of DUET against ICL and Morfessor in the
overall and en$\rightarrow$xx settings.
For en$\rightarrow$xx, the absolute baseline scores are also reported.
$\Delta$ denotes the improvement of DUET over the corresponding baseline.
Win denotes the bootstrap win rate.
\textbf{Bold} values indicate nominal one-sided $p \leq 0.05$ without multiple-comparison correction.
}
\label{tab:bootstrap_puzzle}
\end{table*}

\subsection{Mapping Scheme}
\label{app:mapping_scheme}
Since modern LLMs are pretrained on massive corpora,
evaluating them directly on the original benchmark may introduce data leakage
and confound contextual reasoning with memorized linguistic knowledge.
To reduce direct lexical cues from pretraining,
we transform every low-resource language in the benchmark into an artificial ``unknown'' language
using a deterministic 1-to-1 character mapping.
The mapping preserves all structural regularities
but does not guarantee that the model lacks all relevant prior knowledge.
The substitution scheme is described below:
\begin{itemize}
    \item Each language uses an independent character mapping.
    \item For Latin-script characters, vowels are replaced with vowels and consonants with consonants.
    \item The mapping is 1-to-1 within each language.
    \item Diacritics and combining marks are preserved (e.g., $\texttt{\`a} \mapsto \texttt{\`e}$).
    \item Non-Latin characters are replaced with characters from the same Unicode block.
    \item Punctuation, digits, and whitespace remain unchanged.
    \item The substitution is applied only to the source-language side; English translations remain unchanged.
\end{itemize}

\subsection{Experimental Settings}
\label{app:puzzle_settings}
We use three open-weight LLMs for evaluation:
\textsc{Qwen3.6-27B-FP8},
\textsc{Gemma4-31B-it},
and \textsc{GPT-oss-120B}.
For each model,
generation hyperparameters follow the officially recommended settings,
with a unified maximum generation length of 81,920 tokens
(other hyperparameters are listed in Table~\ref{tab:puzzle_config}).
All experiments are conducted using the \texttt{vLLM} library~\citep{vllm}
on a single NVIDIA H100 GPU.
To account for generation variance,
each method is evaluated over five runs using a fixed set of seeds.
The prompts used for inference and LLMSeg
are provided in Figures~\ref{fig:prompt_puzzle}, \ref{fig:llm_gen_toks}, respectively.
\begin{table}[htbp]
\centering
\resizebox{\columnwidth}{!}{
\begin{tabular}{lccc}
\toprule
Setting & \textsc{Gemma4} & \textsc{GPT-oss} & \textsc{Qwen3.6} \\
\midrule
Temperature      & 1.0  & 1.0          & 0.6 \\
Top-$p$          & 0.95 & 1.0          & 0.95 \\
Top-$k$          & 64   & Disabled     & 20 \\
Presence penalty & 0.0  & 0.0          & 1.5 \\
Quantization     & FP8  & Native MXFP4 & Native FP8 \\
KV-cache dtype   & FP8  & Default      & Default \\
\bottomrule
\end{tabular}
}
\caption{Inference configurations.}
\label{tab:puzzle_config}
\end{table}

\subsection{Statistical Significance Test}
\label{app:stat_test}

We follow \citet{koehn-2004-statistical} and perform a hierarchical bootstrap test,
treating each puzzle as a resampling unit.
All evaluation instances belonging to the same puzzle are kept together.
Because each method is evaluated over five independent runs, we extend
the standard procedure with a second resampling stage. Specifically,
in each bootstrap replicate, we first sample the eligible puzzles with replacement.
For each sampled puzzle and each method,
we then independently sample five runs with replacement from the five observed runs
and average their puzzle-level scores. We aggregate these scores across the sampled puzzles,
weighted by the number of evaluation instances in each puzzle,
and compute the win rate for \textsc{DUET} against the baseline.
We perform a one-sided test,
$H_0:\Delta \leq 0$ vs. $H_1:\Delta > 0$, using
1,000,000 bootstrap replicates, where
$\Delta=\mathrm{score}(\text{DUET})-\mathrm{score}(\mathrm{\text{Baseline}})$.

Because the pseudo benchmark reduces direct lexical cues, we limit this analysis to that version.
The results are reported in Table~\ref{tab:bootstrap_puzzle}.
Additionally, we observe that the benefits of word-segmentation prompting
are primarily concentrated in the en$\rightarrow$xx translation direction.
Accordingly, we also perform significance testing restricted to this subset.
Across the four evaluation metrics, nominally significant improvements are observed primarily in comparisons
between DUET and ICL,
with BLEU-2 and EM reaching this threshold more often than chrF-3 and $1-\mathrm{cTER}$.
These one-sided $p$-values are exploratory and are not corrected for the multiple models, metrics, directions, and baselines in Table~\ref{tab:bootstrap_puzzle}.
DUET also rarely reaches nominal significance over the baselines
when using \textsc{GPT-oss-120B}.
The only such \textsc{GPT-oss-120B} comparisons are against ICL in the en$\rightarrow$xx direction, on BLEU-2 and EM.

\begin{table*}[t]\centering\small
\resizebox{\textwidth}{!}{
\begin{tabular}{lrrrrrrrrr}
\toprule
System & DUET & Morfessor & Linguistica & MorphAGram & BPE & whole-word & every-atom & DeepSeek & Qwen \\
\midrule
Morph.\ F1 & \textbf{.482} & .471 & .467 & .213 & .146 & .447 & .076 & .222 & .253 \\
Bnd.\ F1 & .164 & .167 & .105 & \textbf{.316} & .301 & .000 & .272 & .259 & .112 \\
Oracle@1 & .583 & .572 & \textbf{.587} & .129 & .035 & .580 & .017 & .179 & .284 \\
Oracle@8 & \textbf{.903} & .879 & .587 & .129 & .035 & .580 & .017 & .318 & .476 \\
\bottomrule
\end{tabular}
}
\caption{Strict character-blind test comparison, averaged equally over 35 cells.
LLM values use failure-as-zero scoring; all non-LLM requests succeeded (2,240/2,240).}
\label{tab:llm-strict-direct}
\end{table*}

\begin{table*}[t]\centering\small
\begin{tabular}{lrrrrrr}
\toprule
Model & success/requested & fail & Morph.\ F1 & Bnd.\ F1 & Oracle@1 & Oracle@8 \\
\midrule
DeepSeek-V4-Flash & 1653/2240 & 26.2\% & .306 & .351 & .247 & .428 \\
Qwen3.6-27B & 1693/2240 & 24.4\% & .307 & .180 & .320 & .612 \\
\bottomrule
\end{tabular}
\caption{Strict LLM audit conditional on successful requests.}
\label{tab:llm-strict-conditional}
\end{table*}

\begin{figure*}[t]\centering
\includegraphics[
        width=\textwidth,
        trim=0.0cm 1.0cm 0.0cm 0.0cm,
        clip
        ]{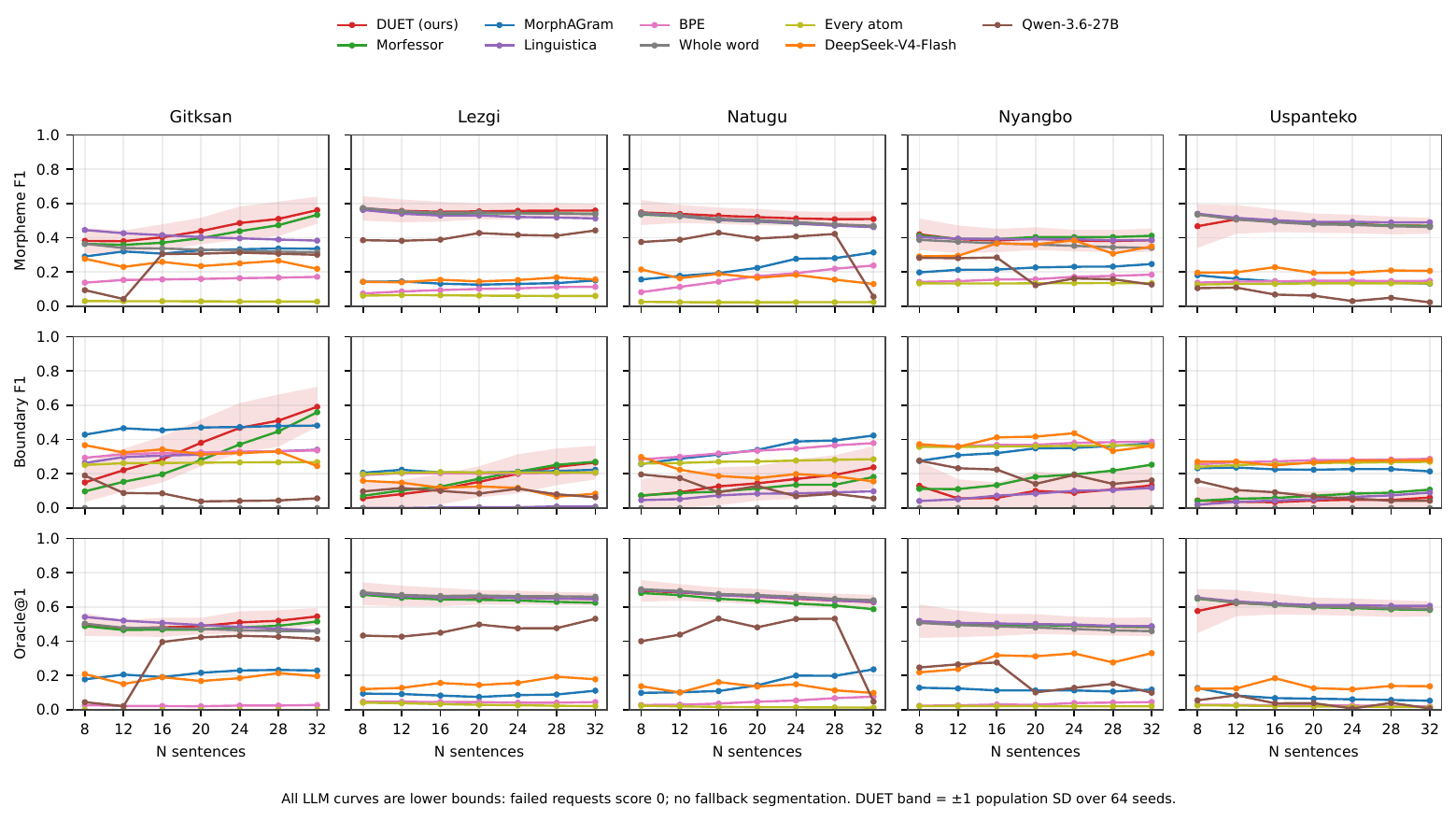}
\caption{Strict character-blind test curves.
LLM curves use failure-as-zero scoring; the DUET band is $\pm1$ population SD over 64 seeds.}
\label{fig:llm-test-comparison}
\end{figure*}

\begin{figure*}[t]\centering
\includegraphics[
        width=\textwidth,
        trim=0.0cm 1.0cm 0.0cm 0.0cm,
        clip
        ]{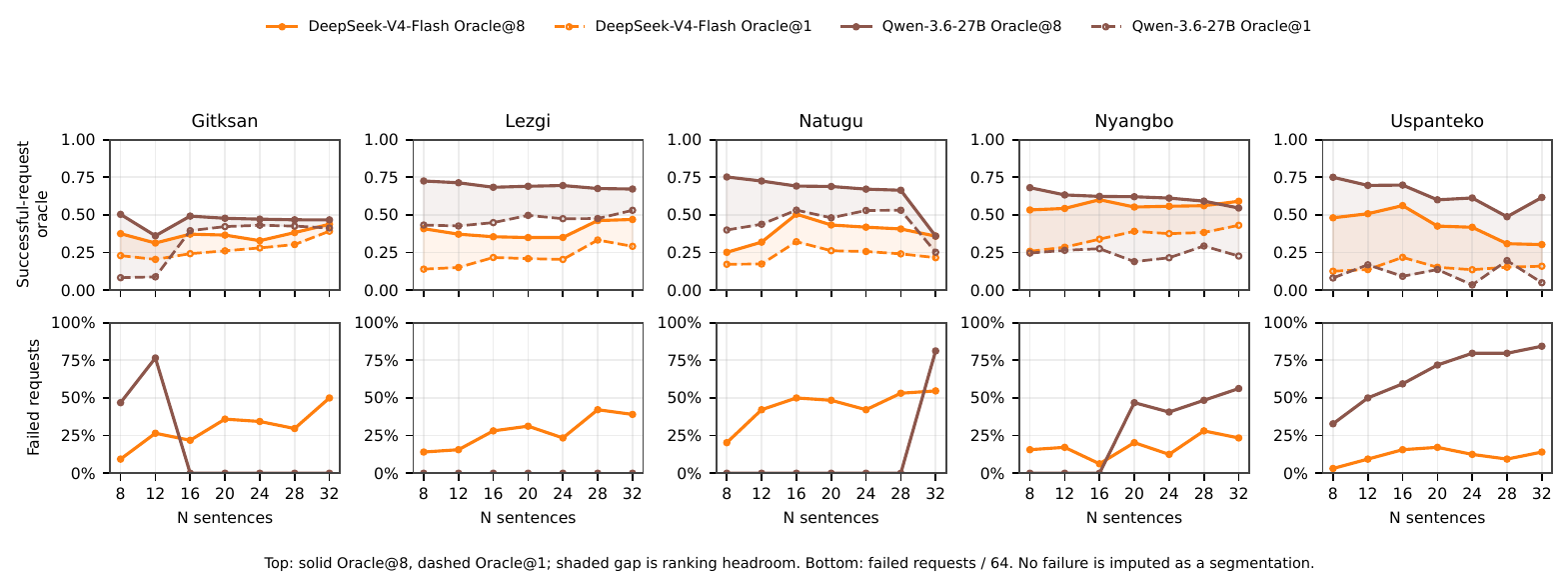}
\caption{Strict LLM audit.
Top: successful-request Oracle@8 (solid) and Oracle@1 (dashed); bottom: request failure rate.}
\label{fig:llm-strict-audit}
\end{figure*}

\section{Zero-Shot LLM Segmentation Audit}
\label{app:llm-audit}

The audit asks whether a zero-shot LLM can produce the segmentation
under exactly the character-blind protocol used for the non-LLM systems.

\subsection{Prompt and Generation Settings}
\label{app:llmprompt}

The zero-shot LLM audit makes one batch request per sampled vocabulary, 
covering every word in that language--$N$--seed cell: no per-word calls and no gold examples.
Before sampling,
each language's test-set characters are replaced by a fixed 1-to-1 mapping.
The mapping scheme is the same as the description in Appendix~\ref{app:mapping_scheme}.
Neither the map nor the language name is sent to the model.
The same template, shown in Figure~\ref{fig:llm-prompt}, 
is used for DeepSeek-V4-Flash~\citep{deepseekv4}
and \textsc{Qwen3.6-27B}, 
requesting up to $K=8$ ranked candidates per word.

\paragraph{Generation settings.} Both models are sampled at temperature $0.15$, top-$p$ $0.9$, top-$k$ $40$, with a 32,768-token output cap and at most eight candidates per word.
A reply is parsed as JSON and every candidate is checked to reconstruct its word exactly, character for character.
If any input word is missing or any word has no valid candidate after checking,
the entire request is recorded as failed and contributes no metric value; no whole-word fallback is inserted.
Results evaluated only on successful requests and on the full set of samples (with failed requests scored as zero)
are reported in Table~\ref{tab:llm-strict-conditional}
and Table~\ref{tab:llm-strict-direct}, respectively.

\subsection{Strict Character-Blind Test Comparison}
\label{app:llmstrict}

We reran every non-LLM baseline on the same character-blind SIGMORPHON 2023 test split used by the audit: Gitksan, Lezgi, Natugu, Nyangbo, and Uspanteko; $N\in\{8,12,16,20,24,28,32\}$;
64 deterministic seeds per language--$N$ cell; and $K=8$.
Per-seed vocabulary, scored-word,
and gold-segmentation records agree exactly between the baseline rerun and both strict LLM caches.
The full test curves are shown in Figure~\ref{fig:llm-test-comparison}.
A more detailed view containing only the LLM curves is shown in Figure~\ref{fig:llm-strict-audit}.
Table~\ref{tab:llm-strict-direct} averages the 35 cells equally and assigns failed LLM requests a zero score;
these are failure-as-zero scores, not scores from fallback segmentations.
Baseline Oracle@8 is recomputed from precisely its first eight raw candidates.

\section{Computational Complexity}
\label{app:complexity}

\subsection{Time and Space Complexity}
\label{app:time_space}

We retain the notation of \S\ref{sec:method}:
$\Vocab$ is the normalized type vocabulary,
$N=|\Vocab|$, $|w|$ is the atom length of word $w$,
and $\mathcal A$ is the atom alphabet.
We additionally define $\ell_{\max}=\max_{w\in\Vocab}|w|$
and $\ell_{\Sigma}=\sum_{w\in\Vocab}|w|$.
For iterative procedures,
$I_{\mathrm{DUET}}$, $I_{\mathrm{Morf}}$, $I_{\mathrm{MAG}}$,
and $I_{\mathrm{BPE}}$ denote,
respectively, the numbers of DUET MDL-guided search sweeps,
Morfessor epochs, MorphAGram sampling sweeps, and BPE merge steps.
These quantities are exposed rather than silently treated as constants.  In our configuration, $I_{\mathrm{MAG}}=1000$
and $I_{\mathrm{BPE}}\leq 2000$;
BPE also stops after ten consecutive merges without a new MDL minimum.

For DUET, let $\rho\leq\ell_{\max}-1$ be the final recursive paradigm level.
At level $\lambda$, let $\sigma_\lambda=|\mathcal S_\lambda|$
and $\varphi_\lambda=|\mathcal F_\lambda|$ be the numbers of distinct stems and frames,
$\eta_\lambda=|\mathcal E_\lambda|$ the number of edges in the pruned frame--stem graph,
$\omega_\lambda$ the number of frame/stem membership incidences retained by accepted paradigms,
and $\xi_\lambda$ the work required to project and propagate block
occurrences.
Let $\Gamma=\sum_{w\in\Vocab}|\mathrm{Cand}(w)|$ be the total number of structural segmentation candidates.
The already-defined $K$ remains the output candidate budget,
and $I_{\mathrm{tree}}\leq50$ denotes the epoch count of the conditional construction-tree MDL.
The structural quantities
$\sigma_\lambda,\varphi_\lambda,\eta_\lambda,\omega_\lambda,\xi_\lambda$,
and $\Gamma$ make DUET's bound more informative than a bound expressed only through $N$ and $\ell_{\max}$.
$G$ is the size of the MorphAGram grammar and $\mathcal Z$ is its adaptor
cache.

\begin{table*}[t]
\centering
\small
\setlength{\tabcolsep}{4pt}
\begin{tabularx}{\textwidth}{@{}lXX@{}}
\toprule
System & Time Complexity & Space Complexity \\
\midrule
BPE
& $\displaystyle
   \begin{aligned}
   O(I_{\mathrm{BPE}}\ell_{\Sigma})
   &{}\subseteq O(I_{\mathrm{BPE}}N\ell_{\max})
   \end{aligned}$
& $O(\ell_{\Sigma}+I_{\mathrm{BPE}}\ell_{\max})$
\\[5pt]

Linguistica
& $O(N^2\ell_{\max}+N\ell_{\max}^2)$
& $O(N\ell_{\max}^2)$
\\[5pt]

MorphAGram
& $\displaystyle
   \begin{aligned}
   O\!\left(I_{\mathrm{MAG}}G\!\sum_{w\in\Vocab}|w|^3\right)
   &{}\subseteq O(I_{\mathrm{MAG}}GN\ell_{\max}^3)
   \end{aligned}$
& $O(G\ell_{\max}^2+N\ell_{\max}+|\mathcal Z|)$
\\[5pt]

Morfessor 2.0
& $\displaystyle
   \begin{aligned}
   O\!\left(I_{\mathrm{Morf}}\!\sum_{w\in\Vocab}|w|^3\right)
   &{}\subseteq O(I_{\mathrm{Morf}}N\ell_{\max}^3)
   \end{aligned}$
& $O(N\ell_{\max}^2)$
\\[5pt]

DUET
& $\displaystyle
   \begin{aligned}
   O\!\biggl(&N\ell_{\max}|\mathcal A|+N\ell_{\max}^2\\[-2pt]
     &+\ell_{\max}\!\sum_{w\in\Vocab}|\mathrm{cuts}(w)|^2\\[-2pt]
     &+\sum_{\lambda=0}^{\rho}\!\left[
       \sigma_\lambda\varphi_\lambda(\sigma_\lambda^2+\varphi_\lambda^2)
       +\xi_\lambda\right]\\[-2pt]
     &+I_{\mathrm{DUET}}\Gamma\ell_{\max}
       +\left(I_{\mathrm{tree}}\!\sum_{w\in\Vocab}|w|^3\right)^{\dagger}
       \\[-2pt]
     &+\left(K\!\sum_{w\in\Vocab}|w|^4\right)^{\dagger}\biggr)
   \end{aligned}$
& $\displaystyle
   \begin{aligned}
   O\!\biggl(&N\ell_{\max}^2+\Gamma\ell_{\max}
       +\left(K\ell_{\max}^3\right)^{\dagger}\\[-2pt]
     &+\sum_{\lambda=0}^{\rho}\!\left[
       \eta_\lambda+\ell_{\max}(\sigma_\lambda+\varphi_\lambda)
       +\omega_\lambda\right]\biggr)
   \end{aligned}$
\\
\bottomrule
\end{tabularx}
\caption{
Worst-case complexity of the evaluated implementations.
Iterative bounds are conditional on the displayed iteration counts.
The DUET row includes its conditional second MDL and top-$K$ generation~(i.e., the most expensive execution path).
Terms marked $\dagger$ are incurred only when the second MDL is active;
storage shared with the first stage is counted once in the unmarked terms.
}
\label{tab:complexity}
\end{table*}

\begin{table*}[htbp]
\centering
\small
\setlength{\tabcolsep}{4pt}
\begin{tabular}{@{}lrrrrrrrr@{}}
\toprule
\multicolumn{9}{@{}l}{\textit{(a) Full training vocabulary}}\\
\midrule
System & Gitksan & Lezgi & Nyangbo & Natugu & Uspanteko & Total (s) & Types/s & Peak RSS \\
\midrule
DUET          &  0.57 &   6.91 &   3.50 &    50.88 &    88.18 &  150.0 &      82 & 171\,MB \\
Morfessor     &  0.20 &   2.36 &   1.71 &     5.50 &    10.13 &   19.9 &     615 &  38\,MB \\
Linguistica   &  0.00 &   0.02 &   0.02 &     0.04 &     0.10 &    0.18 & 67{,}994 &  96\,MB \\
MorphAGram    & 57.42 & 621.28 & 380.76 & 1{,}159.46 & 3{,}319.23 & 5{,}538.2 &  2 & n/a \\
BPE           &  0.02 &   0.50 &   0.33 &     1.54 &     3.25 &    5.6 &  2{,}170 &  27\,MB \\
\midrule
\multicolumn{9}{@{}l}{\textit{(b) Extremely low-resource design point --- milliseconds per (language, seed) cell, macro-averaged}}\\
\midrule
System & $N{=}8$ & $12$ & $16$ & $20$ & $24$ & $28$ & $32$ & $N_{32}/N_{8}$ \\
\midrule
\textit{Types} & 46 & 66 & 84 & 101 & 118 & 134 & 151 & 3.4$\times$ \\
DUET          & 88.4 & 138.1 & 185.9 & 236.5 & 290.6 & 343.5 & 353.3 & 4.5 \\
Morfessor     & 29.8 &  41.0 &  56.3 &  73.0 &  93.4 & 116.1 & 115.9 & 4.2 \\
Linguistica   &  0.5 &   0.6 &   0.8 &   1.0 &   1.1 &   1.3 &   1.4 & 3.1 \\
BPE           &  2.1 &   3.1 &   4.4 &   5.5 &   7.0 &   8.9 &  10.3 & 5.0 \\
\bottomrule
\end{tabular}
\caption{
Measured runtime under the setup of \S\ref{app:runtime}.
(a) Full SIGMORPHON 2023 training vocabulary
($140$, $1{,}572$, $1{,}403$, $2{,}456$ and $6{,}668$ types for
Gitksan, Lezgi, Nyangbo, Natugu and Uspanteko; $12{,}239$ in total);
(b) The smallest low-resource budgets, $N\in\{8,12,\dots,32\}$ sentences
$\times$ $64$ seeds $\times$ $5$ languages,
averaged over languages with equal weight.
The $N_{32}/N_8$ ratios use the four languages available at both endpoints; each absolute-time column averages all languages available at that budget.
}
\label{tab:runtime}
\end{table*}

\paragraph{DUET.}
The front end builds $T_f$ and $T_b$, computes branching entropies,
and detects local entropy maxima in
$O(N\ell_{\max}|\mathcal A|+N\ell_{\max}^2)$ time
in the current tuple-based Python implementation.
Each word contributes only one- and two-boundary analyses from
$\mathrm{cuts}(w)$, so frame construction takes
$O(\ell_{\max}|\mathrm{cuts}(w)|^2)$ time including substring creation,
avoiding enumeration of all $2^{|w|-1}$ possible segmentations.

The dominant cost is paradigm induction.
At recursion level $\lambda$, the implementation closes every eligible
pair of stems against all stems and, dually, every eligible pair of frames
against all frames.
Set intersections, closure scans, and evaluations of
Eq.~\ref{eq:paradigmMDL} therefore require
\begin{equation}
O\!\left(
\sigma_\lambda\varphi_\lambda
(\sigma_\lambda^2+\varphi_\lambda^2)
\right)
=
O(
\sigma_\lambda^3\varphi_\lambda
+
\sigma_\lambda\varphi_\lambda^3
).
\label{eq:duet-paradigm-time}
\end{equation}
Entropy pruning usually keeps $\sigma_\lambda$ and $\varphi_\lambda$ small,
and recursion terminates once no shorter paradigm is accepted.
The worst case nevertheless remains polynomial.

For comparison, the paradigm-closure stage can also be bounded using only
$N$ and $\ell_{\max}$.
Every recursive unit is a substring of an input word, so there are at most
$O(N\ell_{\max}^2)$ distinct stems.
A frame is determined by a parent interval and two internal endpoints,
giving at most $O(N\ell_{\max}^4)$ distinct frames,
and there are at most $\ell_{\max}-1$ recursive levels.
Substituting these quantities into
Eq.~\ref{eq:duet-paradigm-time} yields
\begin{equation}
O(N^4\ell_{\max}^{15}).
\label{eq:duet-loose-bound}
\end{equation}
This deliberately loose bound is not intended to predict practical runtime;
the structural bound in Table~\ref{tab:complexity} is more informative.
Likewise, because the first MDL is optimized iteratively,
a bound expressed only in terms of $N$ and $\ell_{\max}$ would require
either a convergence proof or an explicit bound on
$I_{\mathrm{DUET}}$, so we retain this quantity in
Table~\ref{tab:complexity}.

When Eq.~\ref{eq:treeactive} holds, the second MDL evaluates all binary
splits at each recursive node while updating the affected lexicon entries,
giving the implementation bound
$O(I_{\mathrm{tree}}\sum_{w\in\Vocab}|w|^3)$.
Its memoized top-$K$ substring dynamic program contributes at most
$O(K\sum_{w\in\Vocab}|w|^4)$ time and
$O(K\ell_{\max}^3)$ transient space.
These terms are included in Table~\ref{tab:complexity}; omitting them would
describe only the first MDL rather than the complete DUET pipeline.

\paragraph{Comparison with the baselines.}
BPE has the lowest asymptotic cost among the evaluated systems.
Each merge rescans the current symbol sequence, giving
$O(I_{\mathrm{BPE}}\ell_{\Sigma})$ time and linear corpus storage.

Linguistica is also relatively inexpensive.
The evaluated implementation performs a single deterministic pass with no iterative optimization.
Its dominant cost comes from scanning shared prefixes across the vocabulary, giving
$O(N^2\ell_{\max}+N\ell_{\max}^2)$ time and
$O(N\ell_{\max}^2)$ space.

Morfessor and DUET's construction-tree component both optimize recursive segmentations iteratively.
Morfessor avoids DUET's frame--stem closure and therefore has the better worst-case asymptotic bound, although its epoch count $I_{\mathrm{Morf}}$ is convergence-dependent and uncapped in our configuration.
DUET pays additional polynomial cost to recover explicit paradigms, derivation trees, and multiple structurally licensed candidates, so it should not be regarded as asymptotically faster than Morfessor.

Nevertheless, DUET can be faster in the extremely low-resource settings.
If each word has only $O(1)$ entropy peaks, the frame--stem graphs remain sparse, recursion is shallow, and the construction-tree gate is inactive, the realized computation is approximately
$O(N\ell_{\max}^{2})$ for a fixed alphabet and constant $I_{\mathrm{DUET}}$.
This can be lower than Morfessor's implementation-level
$O(I_{\mathrm{Morf}}N\ell_{\max}^{3})$ cost, especially for longer words requiring multiple training epochs.
This is a regime-dependent observation rather than a general asymptotic advantage.

MorphAGram is the computationally heaviest baseline in our experiments.
Each of its $I_{\mathrm{MAG}}$ sampling sweeps reparses every word with a PCFG proposal, whose chart-parsing complexity is
$O(G|w|^3)$ time and
$O(G|w|^2)$ space per word.
With $I_{\mathrm{MAG}}=1000$, repeated parsing dominates DUET's deterministic candidate generation and MDL optimization at the vocabulary sizes considered here.

Thus, our claim is not that DUET has the lowest asymptotic complexity.
Rather, it replaces exponential segmentation enumeration with an entropy-pruned deterministic search whose structural cost is polynomial (conditional on $I_{\mathrm{DUET}}$).
It is asymptotically more expensive than BPE, Linguistica, and Morfessor, but substantially cheaper than the configured 1000-sweep MorphAGram sampler in the extremely low-resource setting,
while additionally producing explicit paradigms and top-$K$ derivations.
When the construction-tree gate in Eq.~\ref{eq:treeactive} is inactive, the
$O(I_{\mathrm{tree}}\sum_{w\in\Vocab}|w|^3)$
and
$O(K\sum_{w\in\Vocab}|w|^4)$
terms are skipped entirely, so realized runtime can be much lower than the global worst-case bound.

\subsection{Runtime Statistics}
\label{app:runtime}

The bounds above are worst-case; this section reports the realized cost.
We time every system in both regimes:
the full training vocabulary (Table~\ref{tab:runtime}a)
and the low-resource sentence budgets the paper actually evaluates
(Table~\ref{tab:runtime}b).

\paragraph{Measurement setup.}
All measurements were taken on one machine with an Intel Xeon E5-2667 v3 @ 3.20\,GHz, Ubuntu 22.04, and Python 3.12.
The timed interval is training plus inference over all word types,
excluding data loading and process start-up;
peak memory is the maximum resident set size of the process tree.
MorphAGram is timed from a full run of its live \texttt{py-cfg} sampler
at $I_{\mathrm{MAG}}=1000$.

\section{AI Assistance Usage}

We used AI-assisted tools, including
ChatGPT\footnote{\url{https://chatgpt.com}}
and Claude\footnote{\url{https://claude.ai}},
to support code development and writing refinement
(in accordance with ACL Policy on AI Writing Assistance).
In addition, we utilized Claude Code and Codex for implementation.

\begin{figure*}[t]
\centering
\begin{tcolorbox}[
    width=\textwidth,
    colback=white,
    colframe=black,
    colbacktitle=gray!20,
    coltitle=black,
    title=\textbf{Prompt Template for Linguistic Reasoning},
    fonttitle=\bfseries,
    boxrule=0.6pt,
    arc=2mm,
    left=2mm,
    right=2mm,
    top=1mm,
    bottom=1mm,
    fontupper=\small\ttfamily
]

{\bfseries\color{blue!70!black}System}

\vspace{0.3em}
\hrule
\vspace{0.5em}

You are a linguistics expert.
Translate between the specified languages using the provided context.
Return only a valid JSON array as the final output.

\vspace{0.6em}

{\bfseries\color{blue!70!black}User}

\vspace{0.3em}
\hrule
\vspace{0.5em}

\textbf{Source language:}

\texttt{<SOURCE\_LANG>}

\vspace{0.4em}

\textbf{Target language:}

\texttt{<TARGET\_LANG>}

\vspace{0.4em}

\textbf{Metadata:}

\texttt{<META>}

\vspace{0.6em}

\textbf{Context Examples}

\texttt{<CONTEXT\_LINES>}

%%%%%%%%%%%%%%%%%%%%%%%%%%%%%%%%%%%%%%%%%%%%%%%%%%%%%%%%%%%%%%%
%% Only for N-best Tokenization
%%%%%%%%%%%%%%%%%%%%%%%%%%%%%%%%%%%%%%%%%%%%%%%%%%%%%%%%%%%%%%%

\vspace{0.8em}

\begin{tcolorbox}[
    colback=blue!3,
    colframe=blue!60,
    colbacktitle=blue!10,
    coltitle=black,
    title=\textbf{Only for N-best Segmentation},
    boxrule=0.4pt,
    arc=1mm,
    left=1mm,
    right=1mm,
    top=0.5mm,
    bottom=0.5mm,
    fontupper=\small\ttfamily
]

Use the following word Segmentations on the source-language side as
reference. They are provided as candidate segmentations and are not
guaranteed to be ground truth.

\vspace{0.4em}

\textbf{Candidate Segmentations (Source Side)}

\texttt{<WORD\_SEGMENTATION\_CANDIDATE\_TABLE>}

\end{tcolorbox}

%%%%%%%%%%%%%%%%%%%%%%%%%%%%%%%%%%%%%%%%%%%%%%%%%%%%%%%%%%%%%%%

\vspace{0.8em}

\textbf{Test Queries}

\vspace{0.4em}

\underline{Translation Direction}

\texttt{<SOURCE\_LANG>} $\rightarrow$ \texttt{<TARGET\_LANG>}

\texttt{<FORWARD\_TEST\_LINES>}

\vspace{0.6em}

\texttt{<TARGET\_LANG>} $\rightarrow$ \texttt{<SOURCE\_LANG>}

\texttt{<REVERSE\_TEST\_LINES>}

\vspace{0.8em}

\textbf{Output Format}

\begin{verbatim}
[
  {"index": 1, "translation": "..."},
  {"index": 2, "translation": "..."}
]
\end{verbatim}

Return one JSON object for each test query, preserving the original
1-based indices and query order.

\end{tcolorbox}

\caption{
Prompt template used in our experiments for linguistic reasoning.
The blue boxed part is included only in the N-best segmentation prompting
setting and omitted for the standard ICL baseline.
}
\label{fig:prompt_puzzle}
\end{figure*}

\begin{figure*}[t]
\centering
\begin{tcolorbox}[
    width=\textwidth,
    colback=white,
    colframe=black,
    colbacktitle=gray!20,
    coltitle=black,
    title={Prompt Template for N-best Segmentation Generation},
    fonttitle=\bfseries,
    boxrule=0.6pt,
    arc=1.5mm,
    left=2mm,
    right=2mm,
    top=1.5mm,
    bottom=1.5mm,
    fontupper=\small\ttfamily
]

%==================================================
% System message
%==================================================

{\bfseries\color{blue!70!black}System}

\vspace{0.3em}
\hrule
\vspace{0.5em}

You are a linguistics expert. Infer possible source-language word
segmentations from a word list only. Return only the requested JSON
object as the final answer.

\vspace{0.8em}

%==================================================
% User message
%==================================================

{\bfseries\color{blue!70!black}User}

\vspace{0.3em}
\hrule
\vspace{0.5em}

{\bfseries\#\# Task}

\vspace{0.3em}

Generate up to
{\texttt{<CANDIDATE\_NUM>}}
possible segmentations for each listed source-language word.
A segmentation splits one surface word into morpheme-like pieces
separated by spaces.

\vspace{0.7em}

{\bfseries Important constraints:}

\vspace{0.3em}

\begin{minipage}{0.97\linewidth}
\ttfamily
-- Only segment the exact source-language words listed below.\\
-- Preserve the original characters and capitalization of each word.\\
-- For every candidate, removing spaces from the segmentation must
   exactly recover the original word.\\
-- Include the unchanged whole word as a fallback candidate when useful.\\
-- Rank the most plausible surface-pattern candidate first.\\
-- Do not translate, gloss, or label morphemes inside the segmentation.\\
-- Return only valid JSON, no markdown and no commentary.
\end{minipage}

\vspace{0.8em}

{\bfseries\#\# Source-language words to tokenize}

\vspace{0.4em}

{\texttt{<WORD\_LIST>}}

\vspace{0.8em}

{\bfseries\#\# Required JSON schema}

\vspace{0.4em}

\begin{tcolorbox}[
    colback=gray!5,
    colframe=gray!55,
    boxrule=0.4pt,
    arc=1mm,
    left=1.5mm,
    right=1.5mm,
    top=1mm,
    bottom=1mm
]
\ttfamily
\{\\
\hspace*{1em}"words": [\\
\hspace*{2em}\{\\
\hspace*{3em}"word": "surface\_form",\\
\hspace*{3em}"candidates": [\\
\hspace*{4em}\{\\
\hspace*{5em}"rank": 1,\\
\hspace*{5em}"segmentation": "surface form",\\
\hspace*{5em}"reason": "short surface-pattern evidence"\\
\hspace*{4em}\}\\
\hspace*{3em}]\\
\hspace*{2em}\}\\
\hspace*{1em}]\\
\}
\end{tcolorbox}

\vspace{0.6em}

Each word should have between 1 and
{\texttt{<CANDIDATE\_NUM>}}
candidates.

\end{tcolorbox}
\caption{
Prompt used to generate word segmentation candidates by LLMs.
}
\label{fig:llm_gen_toks}
\end{figure*}

\begin{figure*}[t]
\centering
\begin{tcolorbox}[
    width=\textwidth,
    colback=white,
    colframe=black,
    colbacktitle=gray!20,
    coltitle=black,
    title={Prompt Template for Zero-shot Morphological Segmentation},
    fonttitle=\bfseries,
    boxrule=0.6pt,
    arc=1.5mm,
    left=2mm,
    right=2mm,
    top=1.5mm,
    bottom=1.5mm,
    fontupper=\small\ttfamily
]

%%%%%%%%%%%%%%%%%%%%%%%%%%%%%%%%%%%%%%%%%%%%%%%%%%%%%%%%%%%%
% System
%%%%%%%%%%%%%%%%%%%%%%%%%%%%%%%%%%%%%%%%%%%%%%%%%%%%%%%%%%%%

{\bfseries\color{blue!70!black}System}

\vspace{0.3em}
\hrule
\vspace{0.5em}

You are an expert morphologist performing SURFACE morphological segmentation over a small word list.
Hard constraints (apply to every word):

\begin{enumerate}
\item Concatenating a candidate's morphemes MUST reproduce that word EXACTLY,
character for character. Do not add, drop, reorder, normalize, or translate any character.

\item Segment the surface form only -- do not restore underlying/canonical forms.

\item Up to 8 candidate segmentations per word, ordered most to least likely. A single-morpheme word is a valid candidate: ["word"].

Strategy: first skim the whole list ONCE to spot recurring prefixes/suffixes/roots shared across words -- briefly, a few lines at most. Then apply that understanding to give each word a concise, direct answer. Do NOT deliberate at length on each word individually -- with hundreds of words in one list, exhaustive per-word reasoning will overrun the response budget and get cut off before every word is covered.

\item Return STRICT JSON of the form \{"segments": {"word1": [["m1", "m2", ...], ...], "word2": [...], ...}\} covering EVERY word in the input list, no prose outside the JSON.
\end{enumerate}

%%%%%%%%%%%%%%%%%%%%%%%%%%%%%%%%%%%%%%%%%%%%%%%%%%%%%%%%%%%%
% User
%%%%%%%%%%%%%%%%%%%%%%%%%%%%%%%%%%%%%%%%%%%%%%%%%%%%%%%%%%%%

\vspace{0.8em}

{\bfseries\color{blue!70!black}User}

\vspace{0.3em}
\hrule
\vspace{0.5em}

Segment every word in this list into surface morphemes:

["word1", "word2", "word3", ... ]

\end{tcolorbox}

\caption{
The exact zero-shot prompt sent to LLMs.
The user message's JSON array is the full character-blinded sampled vocabulary for that cell;
no language identifier or character map is included.
}
\label{fig:llm-prompt}
\end{figure*}

\end{document}